\documentclass{article}
\usepackage{arxiv} 
\usepackage{siunitx} 
\usepackage{graphicx}
\usepackage[square,numbers]{natbib}
\usepackage{amsmath} 

\usepackage[utf8]{inputenc}
\usepackage[english]{babel}

\newcommand{\vol}{{\ooalign{\hfil$V$\hfil\cr\kern0.08em--\hfil\cr}}}
\renewcommand{\headeright}{ }

\usepackage[utf8]{inputenc}
\usepackage[english]{babel}
\usepackage{hyperref}
\hypersetup{
    colorlinks=true,
    linkcolor=red, 
		anchorcolor=black,
		citecolor=green,
    filecolor=cyan,
		runcolor=cyan,
		menucolor=red,
		urlcolor=magenta, 
}

\usepackage[tableposition=top]{caption} 
\usepackage{float}
\usepackage{authblk}
\floatstyle{plaintop}
\restylefloat{table}
\usepackage{lmodern} 
\usepackage{xcolor} 
\usepackage{subcaption}

\title{Squeeze flow of micro-droplets: convolutional neural network with trainable and tunable refinement}

\author[$\dagger$$\star$]{Aryan Mehboudi} 
\author[$\dagger$]{Shrawan Singhal} 
\author[$\dagger$]{S.V. Sreenivasan}

\affil[$\dagger$]{NASCENT Engineering Research Center,
The University of Texas at Austin, Austin, Texas 78758, United States}
\affil[$\star$]{Email: aryan.mehboudi@austin.utexas.edu}

\date{Nov. 15, 2022}

\begin{document}
\clearpage
\maketitle
\thispagestyle{empty}

\begin{abstract}
We propose a platform based on neural networks to solve the image-to-image translation problem in the context of \textit{squeeze flow of micro-droplets}.
In the first part of this paper, 
we present the governing partial differential equations to
lay out the underlying physics of the problem.
We also discuss our developed Python package, 
\href{https://github.com/sqflow/sqflow}{\textit{sqflow}}, 
which can potentially serve as free, flexible, and scalable standardized benchmarks in the fields of machine learning and computer vision.
In the second part of this paper,
we introduce a residual convolutional neural network to solve the corresponding inverse problem: 
to translate 
a high-resolution (HR) imprint image with a specific liquid film thickness to
a low-resolution (LR) droplet pattern image capable of producing the given imprint image for an appropriate spread time of droplets.
We propose a neural network architecture that learns to systematically tune the refinement level of 
its residual convolutional blocks
by using the function approximators that are trained to map a given input parameter (film thickness) to an appropriate refinement level indicator.
We use multiple stacks of convolutional layers the output of which is translated 
according to the refinement level indicators provided by 
the directly-connected function approximators.
Together with a non-linear activation function, such a translation mechanism enables 
the HR imprint image to be refined sequentially in multiple steps 
until the target LR droplet pattern image is revealed.
The proposed platform can be potentially applied to data compression and data encryption.
The developed package and datasets are publicly available on GitHub at \url{https://github.com/sqflow/sqflow}.
\end{abstract}

\newpage
{
  \hypersetup{
    linkcolor=black, 
	}
  \tableofcontents
}

\section{Introduction}
\subsection{Imprint lithography}

The imprint lithography and particularly a subclass of it called step and flash imprint lithography (SFIL) \cite{sreenivasan_nanoimprint_2017}, are considered as promising alternatives to photolithography process. 
In this method, the imprint resist droplets are dispensed on a substrate, followed by imprinting the droplets using a superstrate (template), and curing the resulted liquid film into a desired pattern \cite{reddy_simulation_2005, sreenivasan_nanoimprint_2017}.

\begin{figure}[b!]
\centering
\includegraphics[width=0.75\linewidth]{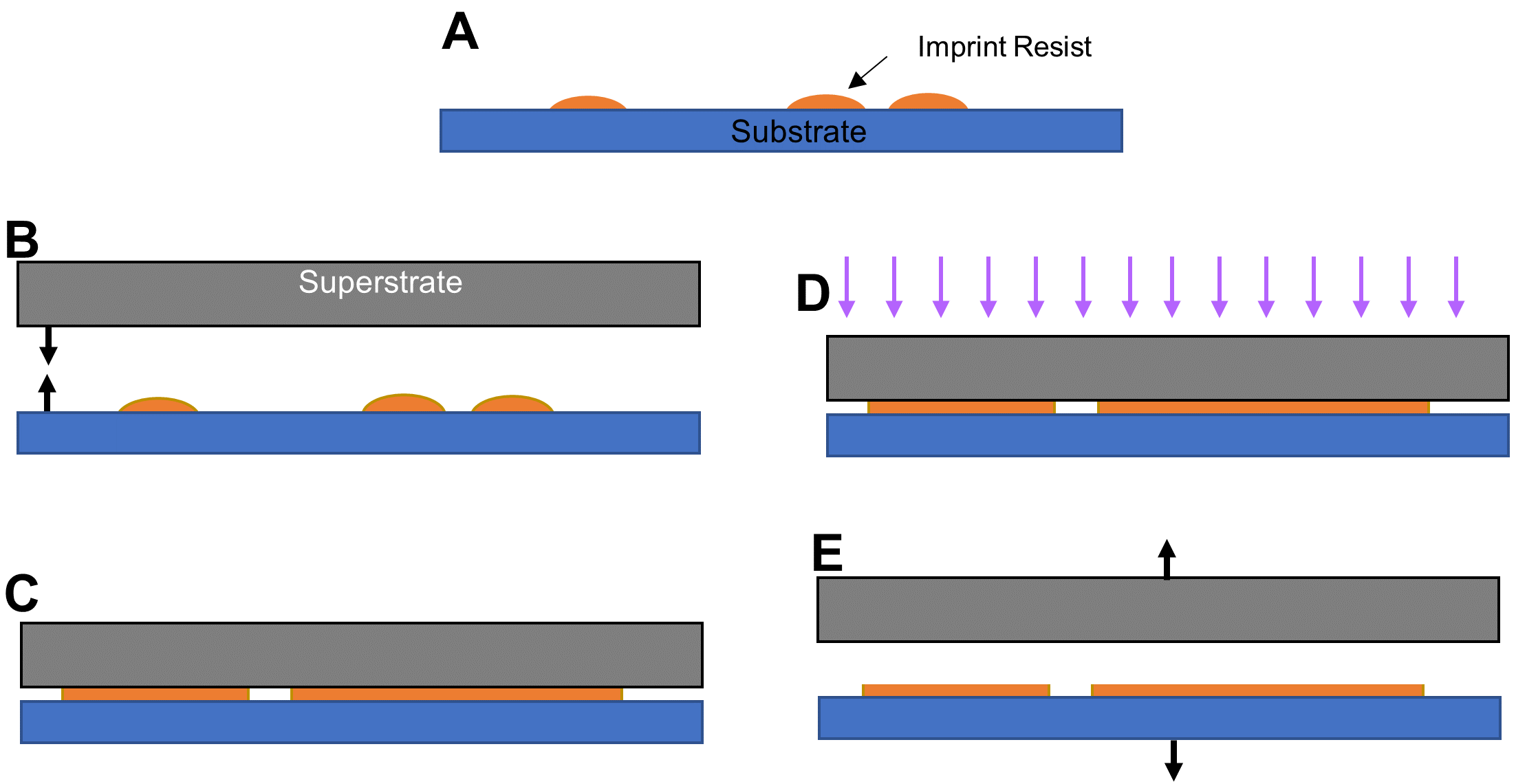}
\caption{A schematic representation of the 
imprint lithography with blank superstrate: 
(A) dispensing imprint resist droplets according to a droplet pattern, 
(B) alignment of superstrate (template) and substrate as needed
followed by squeezing the droplets, 
(C) spread and merge of imprint resist droplets, 
(D) polymerization of imprint resist using the UV exposure, and 
(E) separation of template from substrate.
}
\label{fig-schem_sfil}
\end{figure}

Inspired by SFIL, we aim at developing a methodology for producing a target imprint image from an appropriate droplet pattern map by using a blank superstrate.
Let us consider a pair of parallel plates (substrate and superstrate) initially separated from each other as shown in Fig.~\ref{fig-schem_sfil}.
First, one or multiple droplets of imprint resist are dispensed on the substrate using an inkjet dispenser according to a droplet pattern map.
The superstrate is then brought in contact with the liquid droplets.
Because of the capillary force, the droplets start to spread and merge.
This flow of liquid continues for a period of time, the length of which is called \textit{spread time}.
The resulted liquid film is then polymerized by using an ultraviolet (UV) exposure.
The superstrate is debonded from the substrate at the end.
A schematic representation of these steps can be found in Fig.~\ref{fig-schem_sfil}.
The main difference of this methodology from SFIL is that in this work, we consider a blank superstrate aiming at finding the appropriate initial position of droplets that can produce a given pattern with a specific film thickness.

\subsection{Droplet pattern image \& Imprint image}
In practice, an inkjet printhead has an array of nozzles separated by a finite distance, which is typically of the order of tens of microns.
The pitch of the nozzle array dictates the pixel size of the droplet pattern image.
The droplet pattern image is simply a representation of the firing nozzles.
For each nozzle that dispenses a droplet, the corresponding pixel is considered to have the value of `1' (On).
Otherwise, a value of `0' (Off) is assigned to the corresponding pixel.

On the other hand, the imprint image is a top-down view of the liquid film.
The areas with liquid are assigned a value of `1' (Wet), while a value of `0' is considered for areas without liquid (Dry).
We have developed a physics-based model to simulate the squeeze flow of droplets for any given droplet pattern (forward problem).
The physical domain of interest is discretized into a finite number of computational cells.
Therefore, the pixel size of the imprint image is dictated by the size of the computational cells.

\section{Motivation}
\begin{figure}[tb!]
\centering
\includegraphics[width=0.9\linewidth]{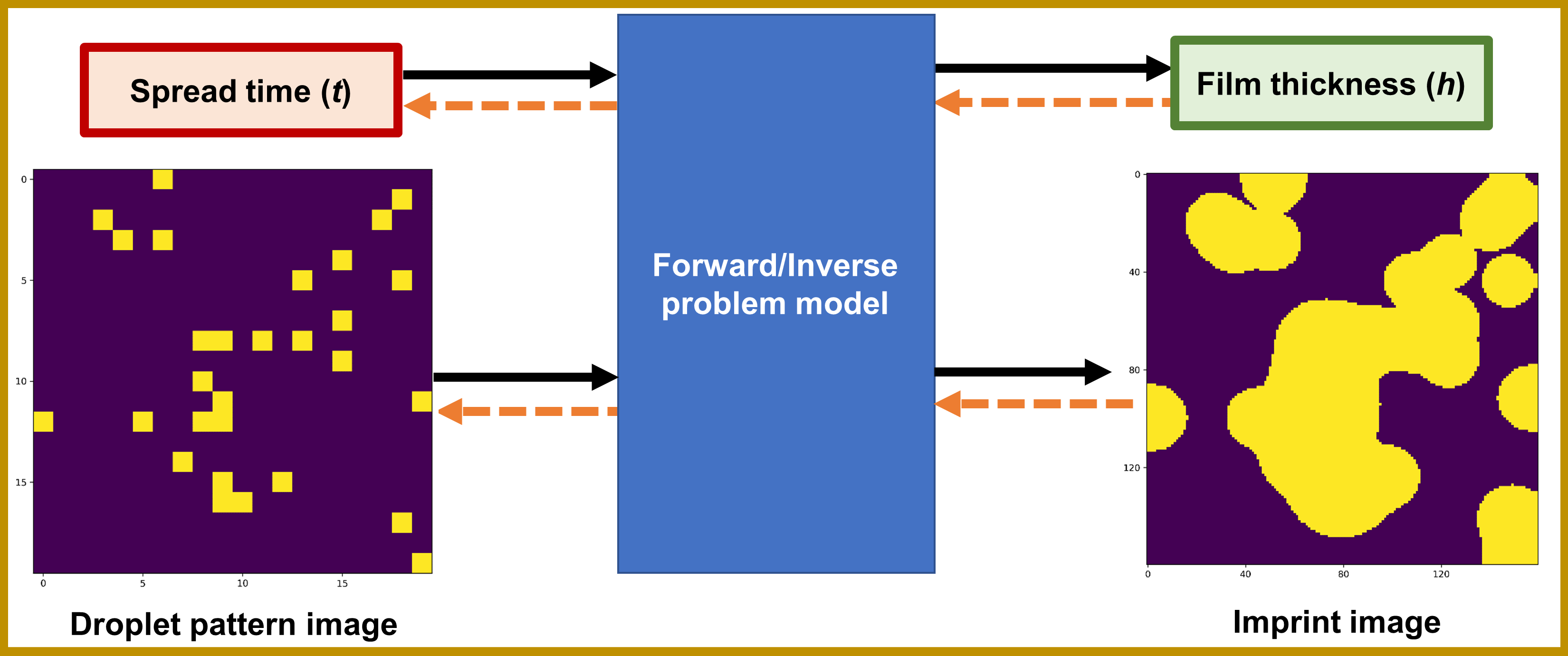}
\caption{
A general schematic representation of the forward (solid black arrows) and inverse (orange dashed arrows) problems for the squeeze flow of droplets.
}
\label{fig-schem_fwd_bwd}
\end{figure}

The schematic representation depicted in Fig.~\ref{fig-schem_fwd_bwd}
illustrates the forward and inverse problems of the squeeze flow of droplets.
Although the forward problem, \textit{i.e.} prediction of the 1. imprint image and 2. film thickness from a given tuple of 1. drop pattern image and 2. spread time, is achievable using a physics-based solver, the inverse problem may not be straightforward.
That is, given an imprint image with a specific film thickness, it may not be trivial to find the appropriate drop pattern image and the required spread time.

\begin{figure}[b!]
\centering
\includegraphics[width=0.9\linewidth]{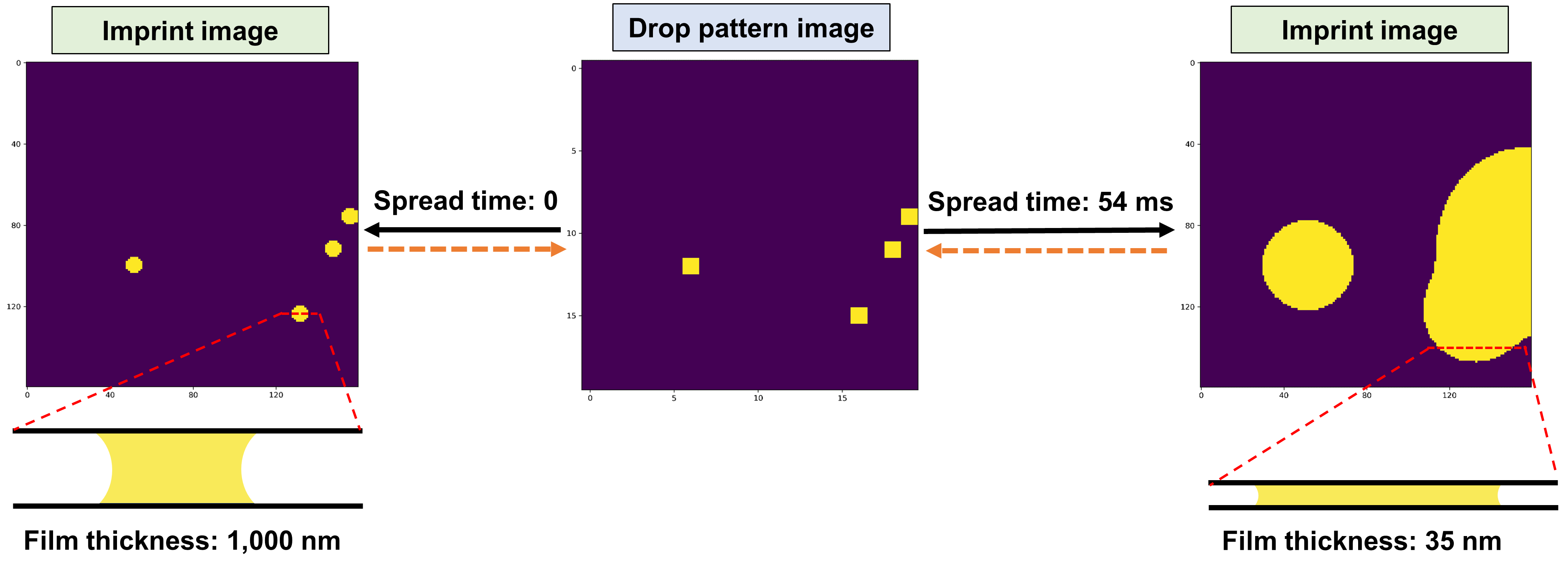}
\caption{A droplet pattern image together with the (numerically) produced imprint images for two different spread times as well as the schematic representations of the cross section view of the liquid film.
The solid black arrows show the forward problem, while the orange dashed arrows denote the inverse problem.
Here, the droplet pattern image has $20\times20$ pixels, while the imprint images consist of $160\times160$ pixels.
The liquid's thermophysical properties and other related parameters are 
discussed in the text.
}
\label{fig-schem_problem}
\end{figure}

As an example shown in Fig.~\ref{fig-schem_problem}, two imprint images are produced with two different film thicknesses for a given drop pattern image under two different spread times.
It can be discerned that the droplet pattern map itself can be a good model representation of the imprint image at the beginning of the spread of droplets (after applying an appropriate transformation to match the dimensions of images). However, as time elapses, the error associated with such a crude model that ignores the film thickness can become noticeably larger.

The goal of this work is primarily to address the inverse problem in the context of squeeze flow of droplets.
However, we believe that providing our compiled dataset as a flexible 
standardized benchmark may potentially benefit the community of machine learning and computer vision for testing algorithms on 
time series images,
image-to-image translation \cite{pang_image--image_2021},
multi-class multi-label classifications \cite{lanchantin_general_2020, yun_re-labeling_2021, liu_emerging_2022}, image super-resolution \cite{wang_deep_2020, li_beginner_2021, liu_blind_2021, saharia_image_2021, maral_single_2022}, 
image deblurring \cite{zhao_new_2020, tao_scale-recurrent_nodate, zhang_deep_2022}, 
etc.
Therefore, we decided to publicly distribute the developed package and the compiled datasets on GitHub at \url{https://github.com/sqflow/sqflow}.

\section{Physics-based solver}
\label{inst_solver}
In this section, we discuss the governing equations for the squeeze flow of droplets.
We also provide a high-level description of the physics-based solver that we have developed for the forward problem.

\subsection{Governing equations}
From the lubrication theory \cite{hamrock_fundamentals_1991}, 
the velocity ($V_l$) and pressure ($p$) fields within a liquid film are coupled as
\begin{equation}
\vec{V_l}=-\frac{h^2}{12\mu_l}\vec{\nabla}p,
\label{eq-Reynolds-Eq}
\end{equation}
where $\mu_l$ denotes the liquid's viscosity, and $h$ shows the instantaneous local 
liquid film thickness.
After change of variable 
$\hat{p}\equiv p-p_{amb}+(\cos\theta_1 + \cos\theta_2)\sigma/h$, 
wherein $\theta_1$ and $\theta_2$ denote the contact angles on the top and bottom plates,
$\sigma$ shows the surface tension, and $p_{amb}$ refers to the ambient pressure,
the governing equation describing the spread of liquid film confined between two plates 
can be written as
\begin{equation}
\begin{split}
\vec{\nabla}.\Big(h^3\vec{\nabla}\hat{p}\Big)&=12\mu_l\frac{\partial h}{\partial t},\\
\text{at interface:}\quad \hat{p}&=0.
\end{split}
\label{eq-gov-02-h}
\end{equation}

The boundary value problem in Eqs.~\ref{eq-gov-02-h} 
needs to be solved for an appropriate instantaneous gap reduction rate $\partial h/\partial t$.
The instantaneous gap reduction rate is determined so that the force balance 
is satisfied as described in \cite{mehboudi_modeling_2022}.

The solver we have developed for the forward problem is based on the finite volume method (FVM).
The volume fraction $f$ is defined for each computational cell as the ratio of liquid volume in the cell to the total volume of that cell. 
After finding the instantaneous pressure and velocity fields, the volume fraction field needs to be modified according to the velocity field to reflect the evolution of the liquid film over time.  
In order to achieve that, the following partial differential equation is solved: 
\begin{equation}
\begin{split}
\frac{\partial f^\ast}{\partial t} + \vec{\nabla}.(f^\ast\vec{V_l})&=0,
\end{split}
\label{eq-vof}
\end{equation}
wherein $f^\ast$ is the modified volume fraction parameter defined as $f^\ast\equiv fh/h_\text{ref}$, and $h_\text{ref}$ is a reference gap distance.
Given the instantaneous velocity and modified volume fraction fields at time $t$, the solver then solves Eq.~\ref{eq-vof} to determine the modified volume fraction field at time $t+\Delta t$. 
The volume fraction ($f$) field can then be obtained for the next time step.
The same process is repeated for each time step until at least one of the termination criteria is met.
A more detailed description of this implementation can be found in \cite{mehboudi_modeling_2022}.

\subsection{Parameters}
The dataset has been compiled for an array of $20\times20$ nozzles with a pitch of $84.5~\mu m$.
We have also set the size of each computational cell to be $1/8$ of the pitch of the nozzle array, \textit{i.e.} $10.5625~\mu m$, which results in imprint images with $160\times160$ pixels.
In addition, we have considered a uniform droplet volume (6 pl) when producing the dataset. 
Some of the other important parameters 
are shown below:

\begin{itemize}
\item{Surface tension: $0.032$ N/m}.
\item{Viscosity: $0.001$ Pa.s}.
\item{Initial thickness of film: $1~\mu m$.}
\end{itemize}

\section{Dataset}
\subsection{Data generation}

The physics-based solver has been used to generate data from different droplet pattern images.
The criteria for terminating each simulation includes:
\begin{itemize}
\item The liquid film covering more than $90\%$ of the field, \textit{i.e.} any imprint image has less than $90\%$ of its pixels \textit{`On'}.
\item The spread time exceeding 1 second, \textit{i.e.} all the examples in the dataset have a spread time ranging from 0 to 1 second.
\item The film thickness decreasing below $5~nm$, \textit{i.e.} all the examples in the dataset have a film thickness ranging from $5~nm$ to $1~\mu m$.
\end{itemize}

Based on the aforementioned termination criteria, each simulation for a given droplet pattern image ($20\times 20$ pixels) produces approximately about 40\textendash60 examples for different spread times.
In addition, the liquid can flow out of the system partially.
In particular, this leads to fewer examples for the cases of having one or a few droplets close to the corner with the coordinates of $(20,20)$.

\begin{figure}[b!]
\centering
\includegraphics[width=0.32\linewidth]{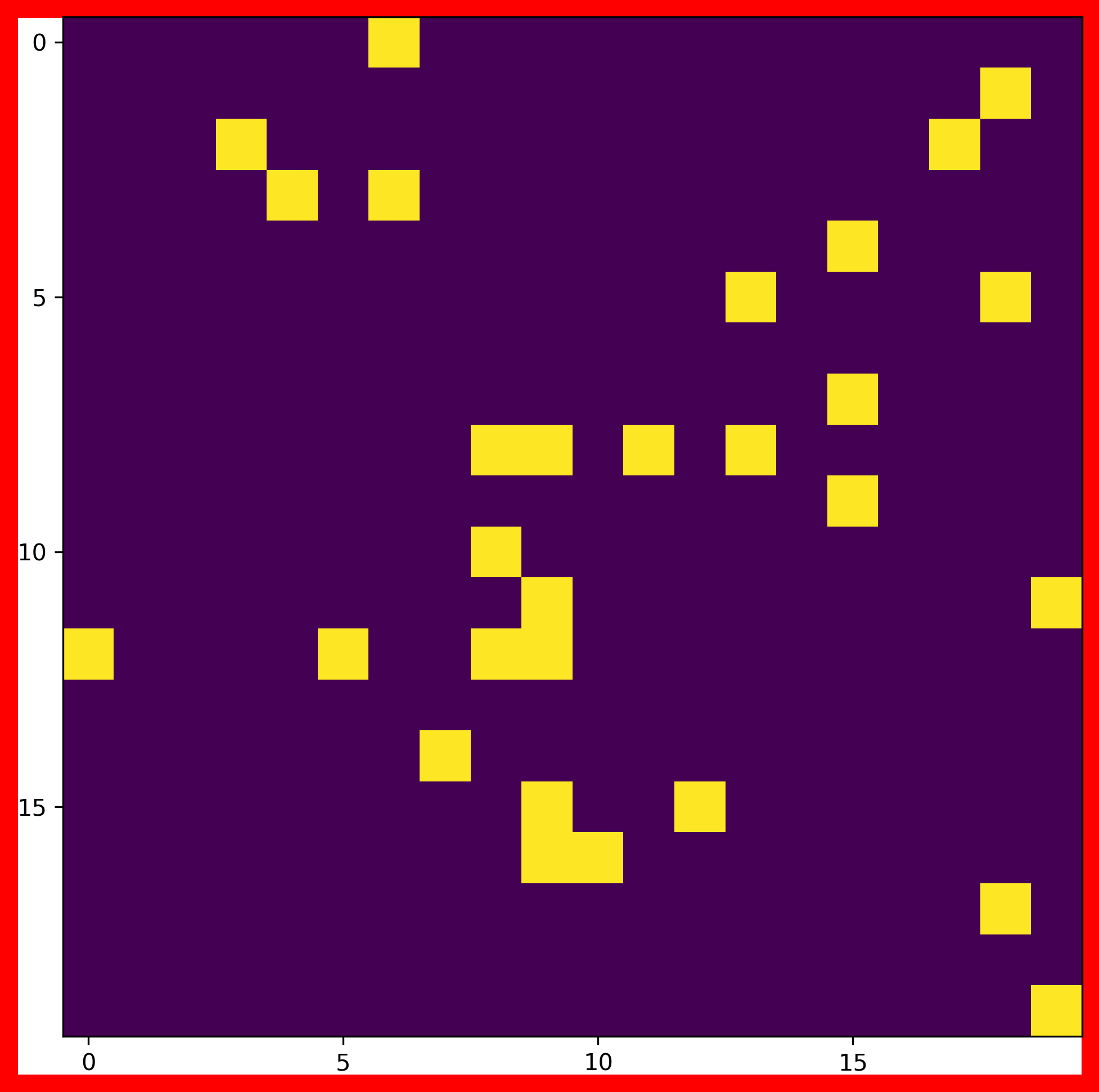}
\includegraphics[width=0.32\linewidth]{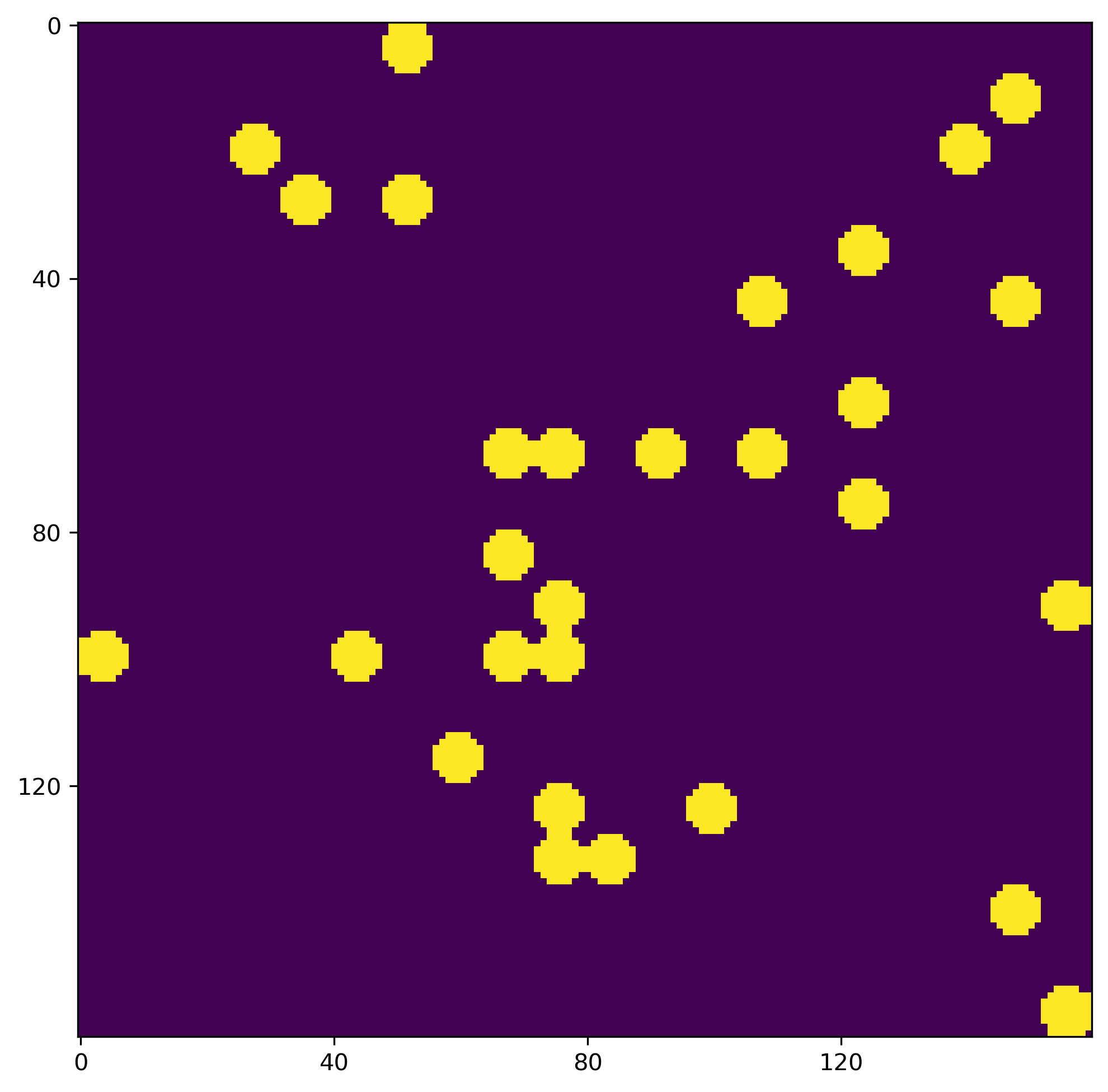}
\includegraphics[width=0.32\linewidth]{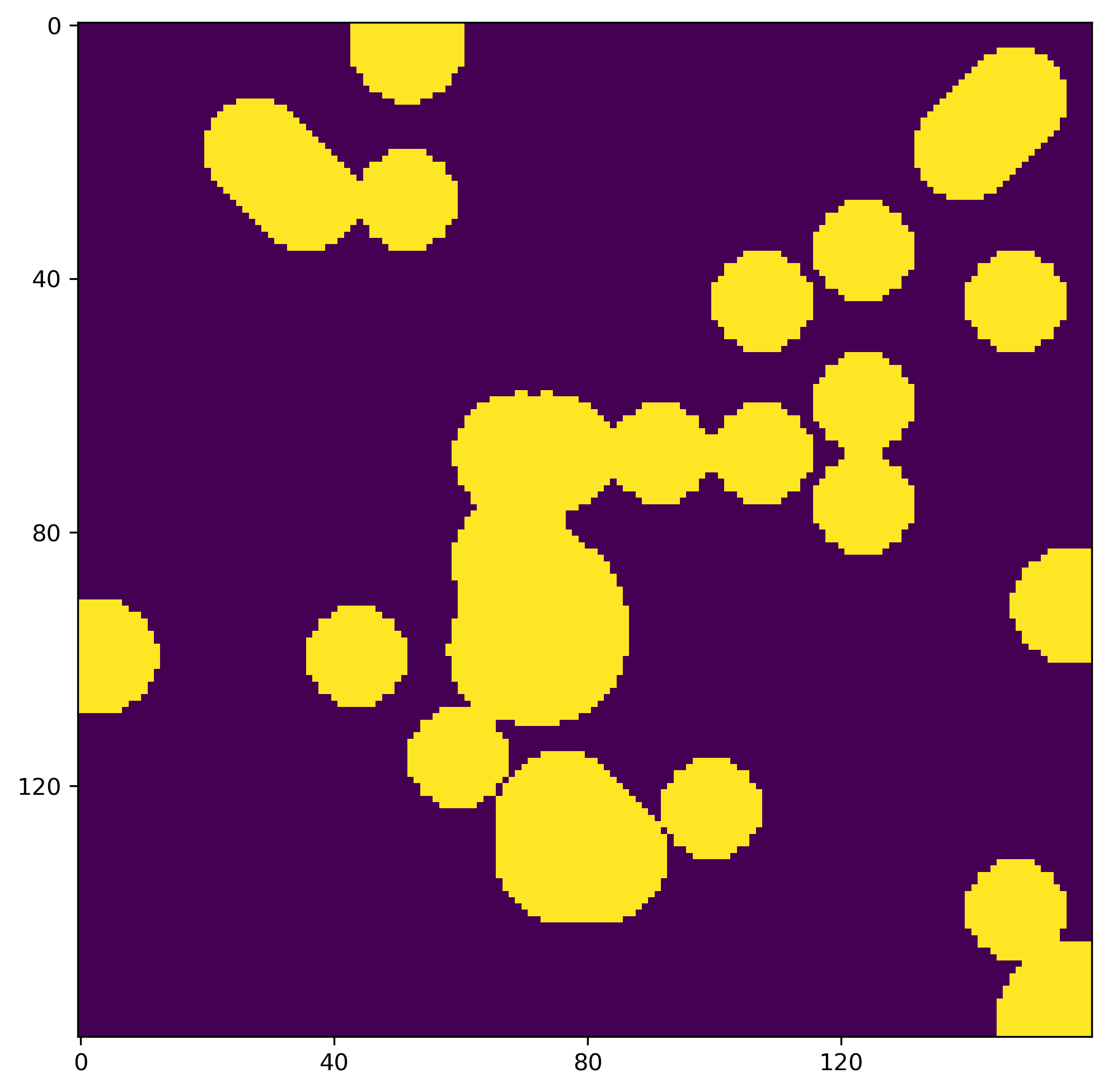}
\includegraphics[width=0.32\linewidth]{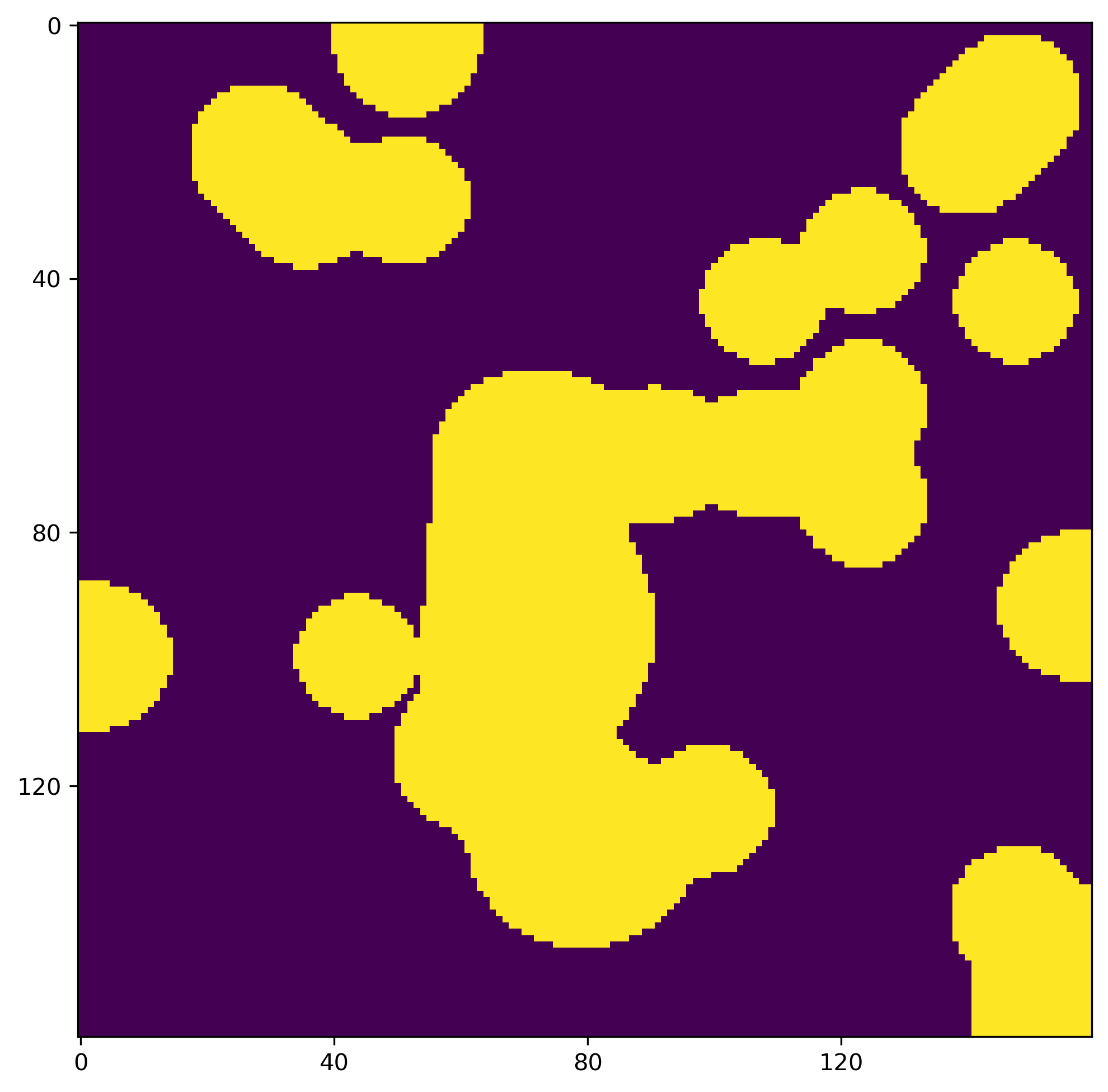}
\includegraphics[width=0.32\linewidth]{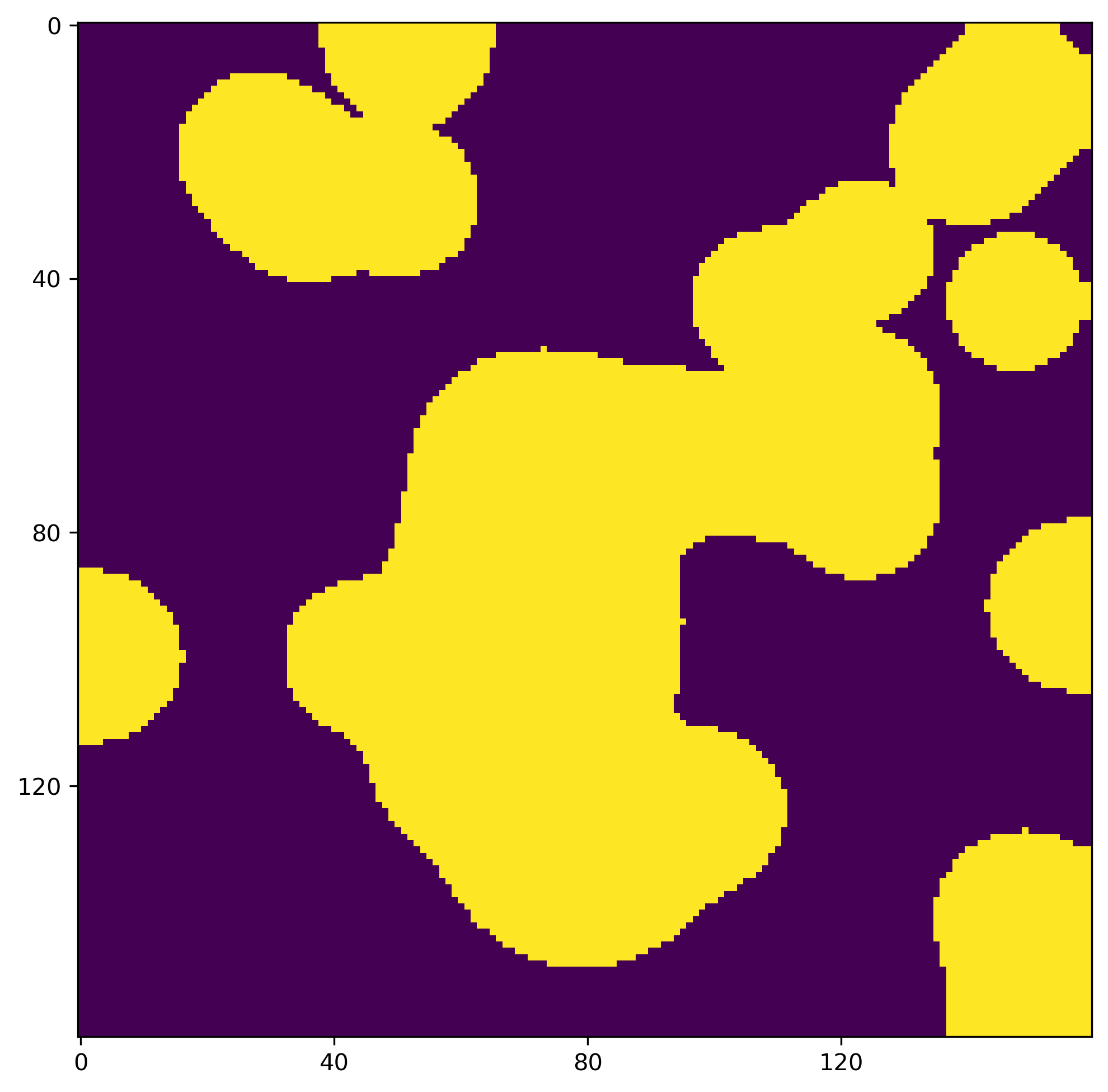}
\includegraphics[width=0.32\linewidth]{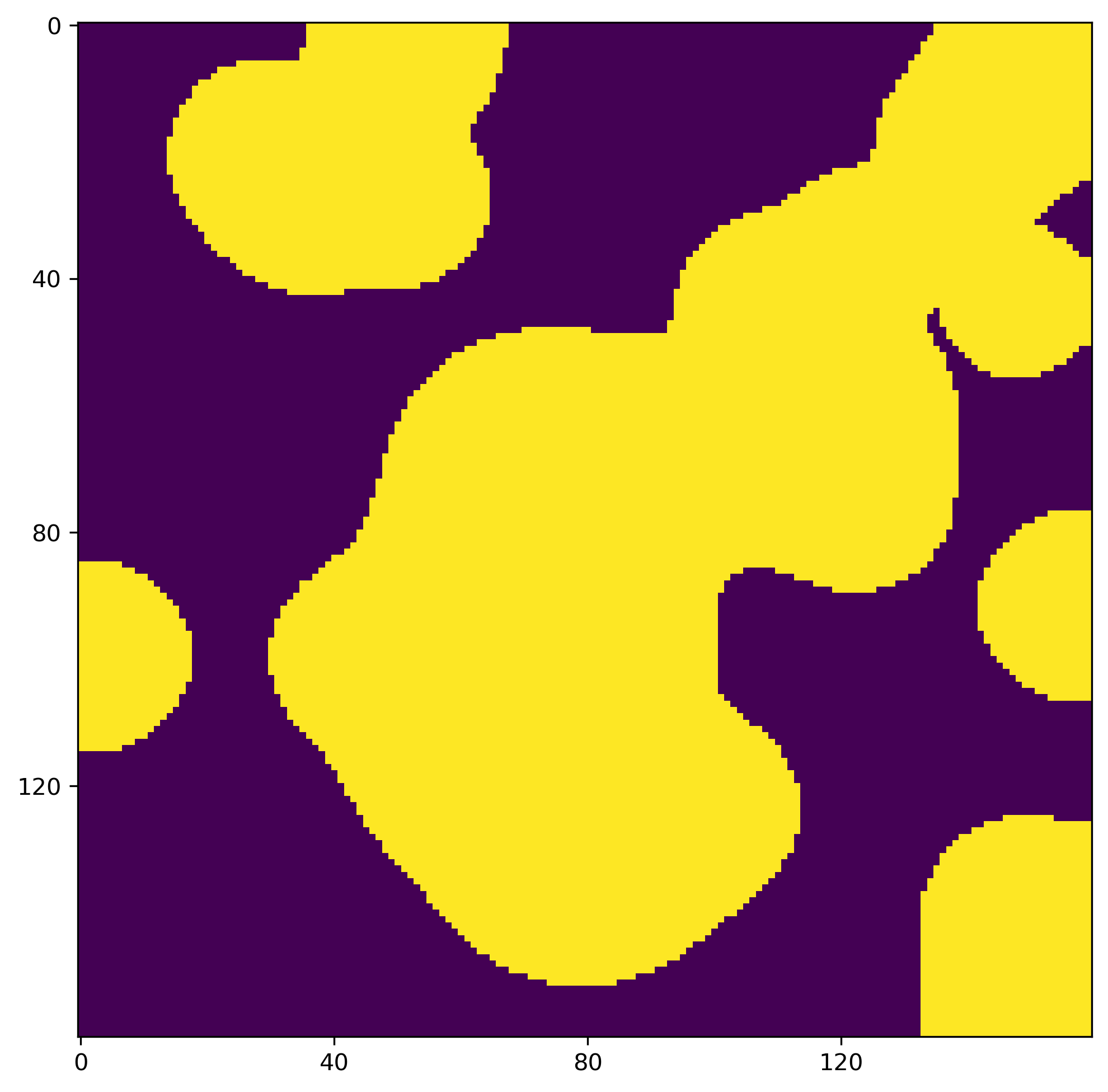}

\caption{
Some of the examples from the dataset: a drop pattern image with $20\times20$ pixels (red border) together with the evolution of the liquid film (imprint images with $160\times160$ pixels) for various spread times:
$0$, $0.6~ms$, $1.8~ms$, $4.2~ms$, and $8.0~ms$, 
increasing from left to right and top to bottom.
The corresponding liquid film thicknesses are
$1~\mu m$, $263~nm$, $179~nm$, $140~nm$, and $116~nm$, 
decreasing from left to right and top to bottom.
}
\label{fig-example_03}
\end{figure}

Some of the examples generated for a random droplet pattern image are shown in Fig.~\ref{fig-example_03}.
We can see that the droplets spread and the film thickness reduces as time elapses.
This phenomenon happens fairly quickly at the beginning, but it slows down over time.

\subsection{Breakdown of the compiled dataset}
We have run the physics-based solver to generate data for several categories based on the 
number of \textit{`On'} pixels in the droplet pattern image, 
\textit{i.e.} number of firing nozzles of the printhead to dispense droplet.
More information about the breakdown of these datasets can be found in Table~\ref{tab-dataset_breadkdown}.

\bgroup
\def\arraystretch{1.5}
\begin{table}[b!]
\centering
\begin{tabular}{|| lcccccccccccc ||}
\hline
Category & 1 & 2 & 3 & 4 & 5 & 7 & 10 & 15 & 20 & 25 & 30 & 40 \\ [0.5ex] 
\hline
\hline
\#Simulations & 400 & 100 & 100 & 100 & 100 & 100 & 100 & 100 & 100 & 100 & 50 & 50 \\[0.5ex] 
\#Examples & 17,047 & 4,912 & 5,627 & 5,882 & 6,189 & 6,441 & 6,203 & 5,745 & 5,101 & 5,028 & 2,295 & 2,007 \\[0.5ex] 
$\overline{|DP|}$ & 1.00 & 2.00 & 3.00 & 4.00 & 4.97 & 6.94 & 9.84 & 14.80 & 19.54 & 24.21 & 29.03 & 38.27 \\[0.5ex] 
\hline
\multicolumn{13}{||l||}{$\overline{|DP|}$: Average count number of \textit{`On'} pixels in the drop pattern images.}\\[0.5ex] 
\hline
\end{tabular}
\caption{Breakdown of the compiled datasets.}
\label{tab-dataset_breadkdown}
\end{table}
\egroup

The dataset for each category is provided independently from others. 
Some of the advantages include:

\begin{itemize}
\item Enabling researchers to select the categories that may best suit their application in future.
\item Providing a convenient way to properly prepare the training, validation, and test datasets 
from the data related to different categories not seen by each other.
\item Reducing the size of dataset by splitting it into smaller partitions.
\end{itemize}

When preparing the training, validation, and test datasets for a supervised learning task, 
one or multiple categories can be used partially or in its entirety.
However, in order to mitigate the data leakage issue,
one should be careful to \textbf{NOT} include 
examples with the same droplet pattern image in more than one dataset.
This can be achieved by using data from different categories when preparing the training, validation, and test datasets.
Alternatively, 
one can preprocess the data to ensure that the examples associated with the same droplet pattern image are not included in more than one of the training, validation, and test datasets.

Fig.~\ref{fig-dataset_breakdown} presents the breakdown of the compiled datasets for different categories in terms of the count number of occurrences for each pixel in the droplet pattern images.
\begin{figure}[t!]
	\centering
	\begin{subfigure}[b]{\textwidth}
			\centering
			\includegraphics[width=0.64\linewidth]{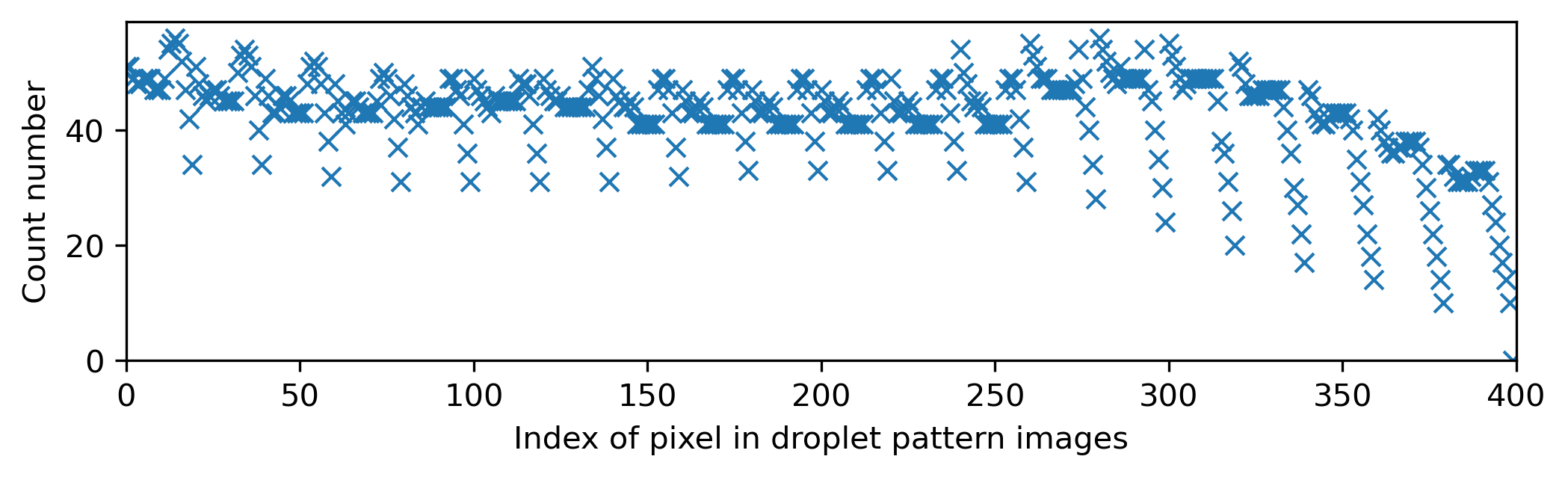}
			\includegraphics[width=0.32\linewidth]{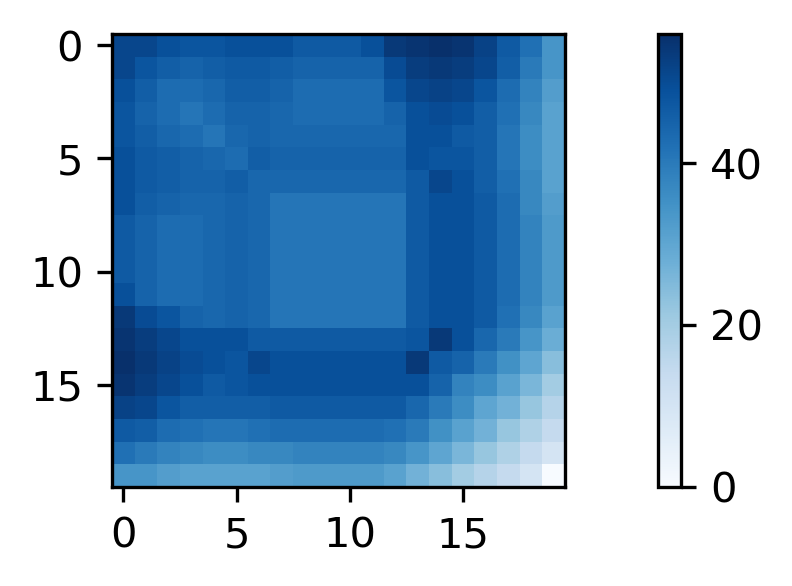}
			\caption{Category: 1}
			\label{cat_01}
	\end{subfigure}
	\hfill
	
	\begin{subfigure}[b]{\textwidth}
			\centering
			\includegraphics[width=0.64\linewidth]{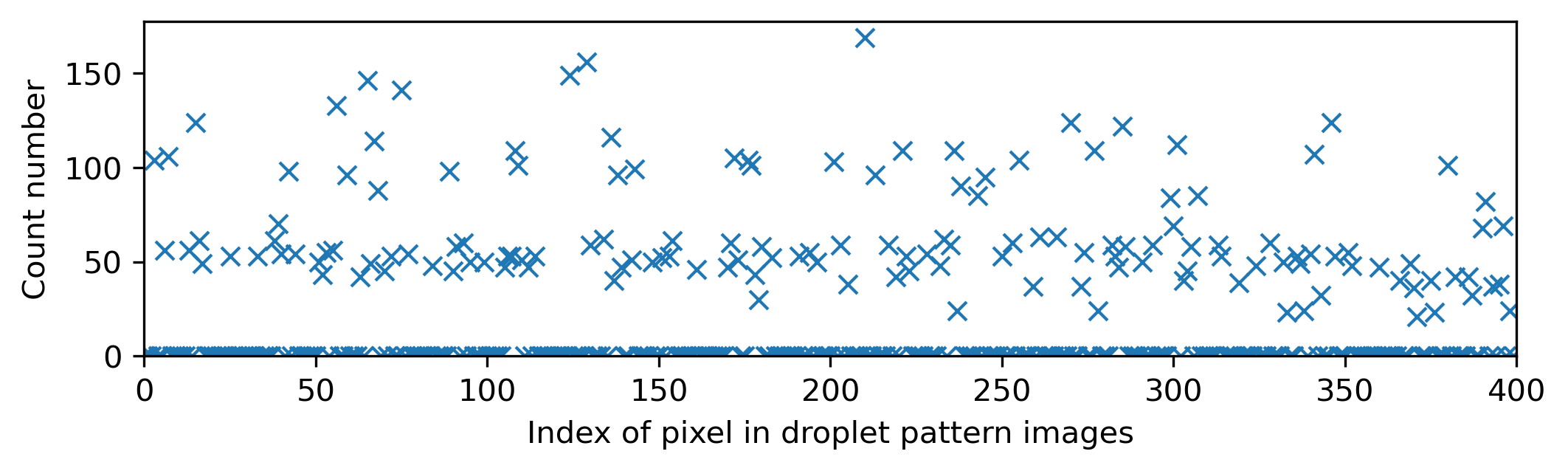}
			\includegraphics[width=0.32\linewidth]{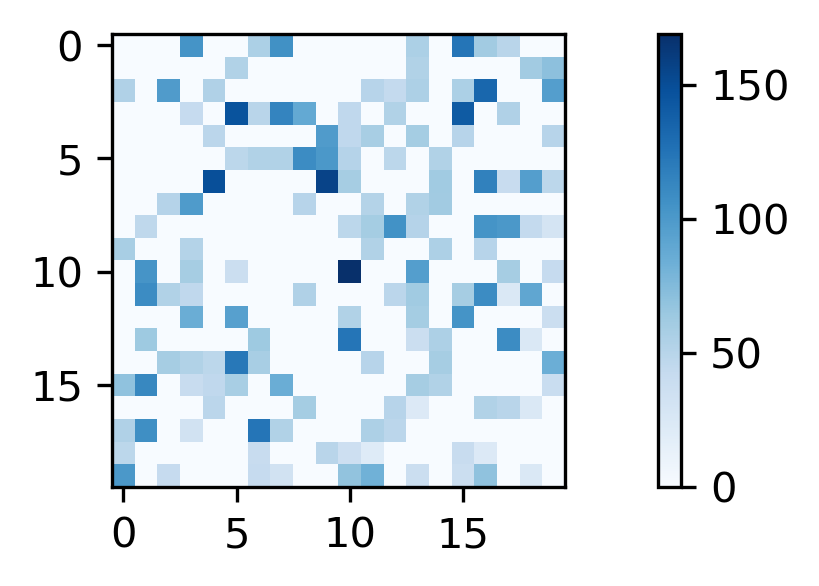}
			\caption{Category: 2}
			\label{cat_02}
	\end{subfigure}
	\hfill

	\begin{subfigure}[b]{\textwidth}
			\centering
			\includegraphics[width=0.64\linewidth]{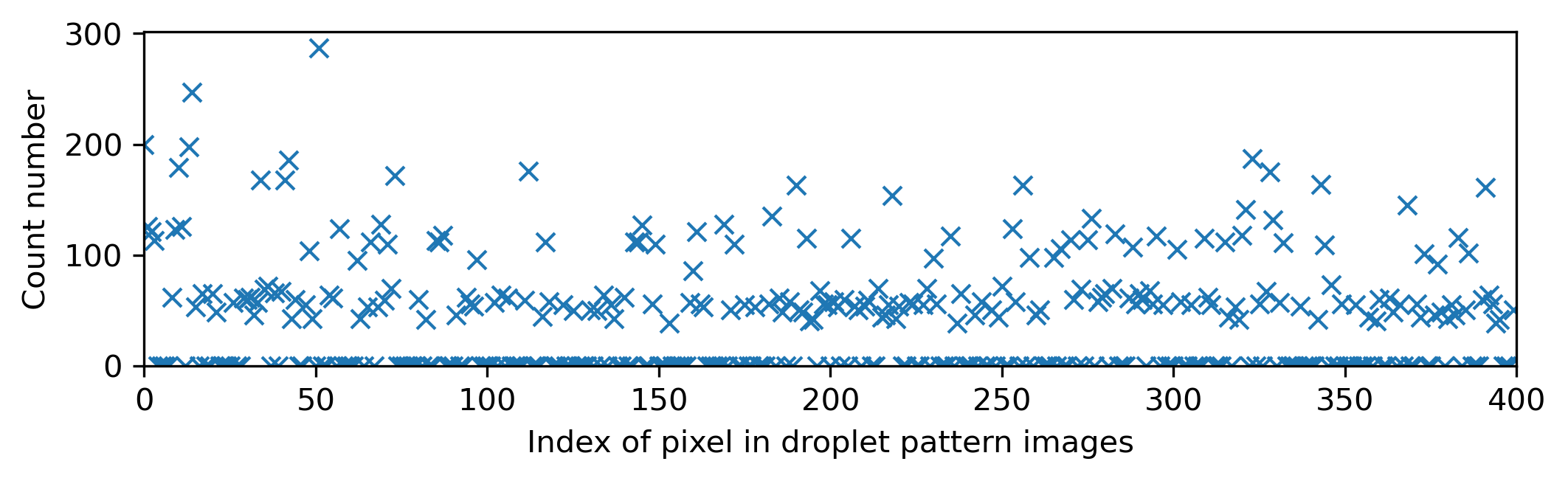}
			\includegraphics[width=0.32\linewidth]{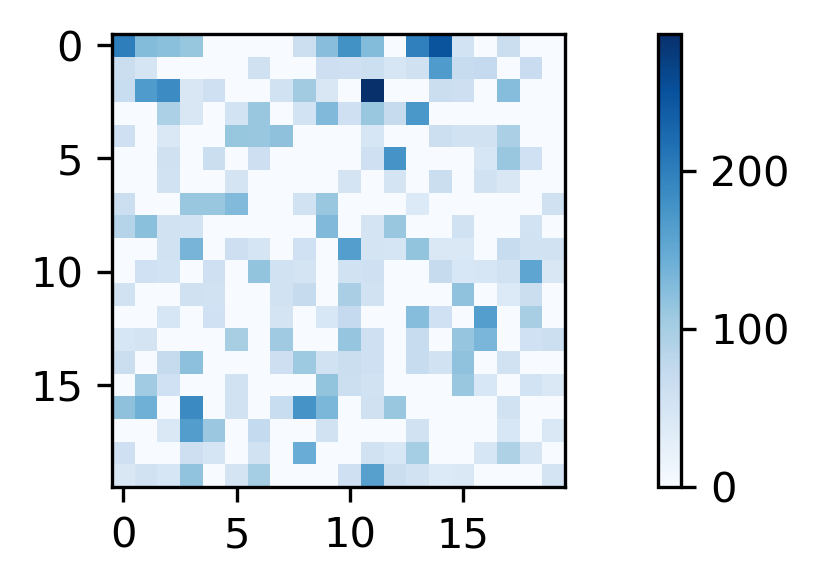}
			\caption{Category: 3}
			\label{cat_03}
	\end{subfigure}
	\hfill
	
	\begin{subfigure}[b]{\textwidth}
			\centering
			\includegraphics[width=0.64\linewidth]{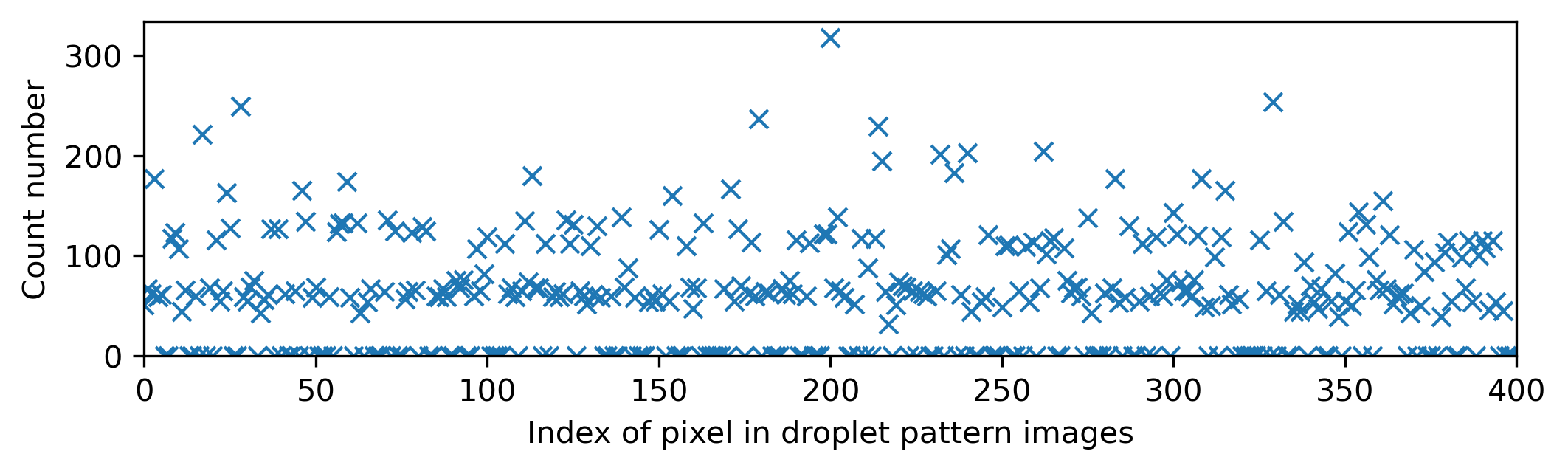}
			\includegraphics[width=0.32\linewidth]{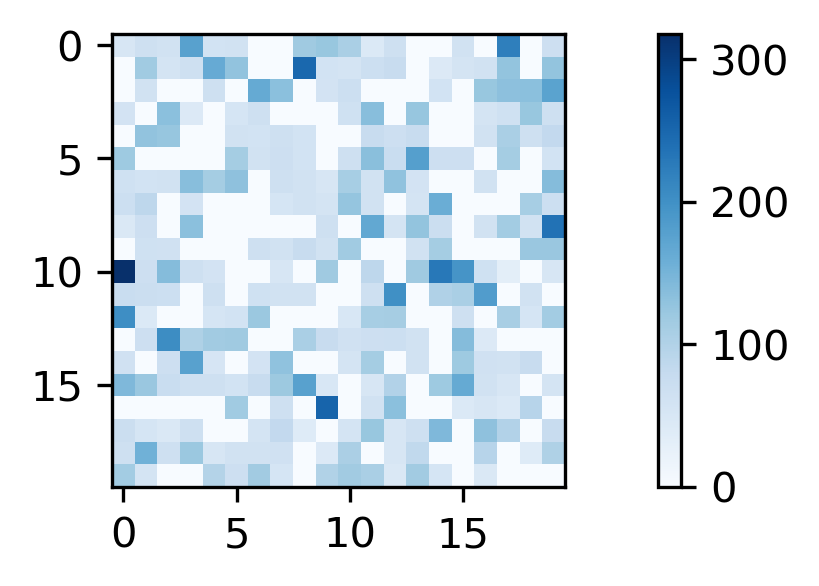}
			\caption{Category: 4}
			\label{cat_04}
	\end{subfigure}
	\hfill
	
\caption{
Breakdown of the compiled datasets: the count number of occurrences for each pixel in the droplet pattern images with $20\times20$ pixels.
}
\label{fig-dataset_breakdown}
\end{figure}

\begin{figure}[!ht]
\ContinuedFloat
\centering

	\begin{subfigure}[b]{\textwidth}
			\centering
			\includegraphics[width=0.64\linewidth]{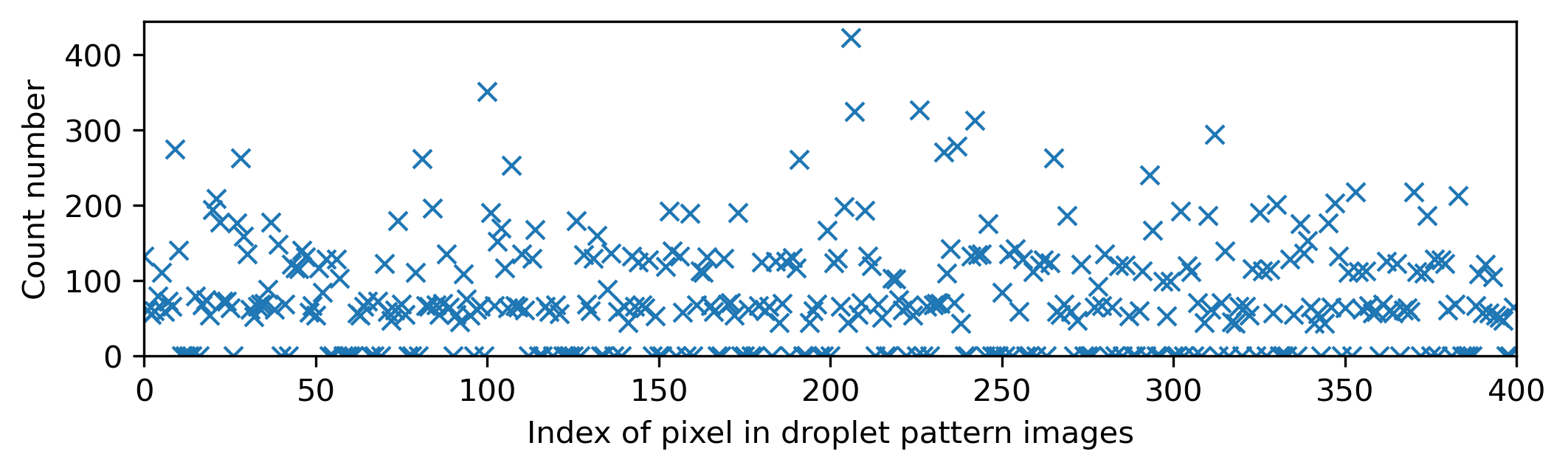}
			\includegraphics[width=0.32\linewidth]{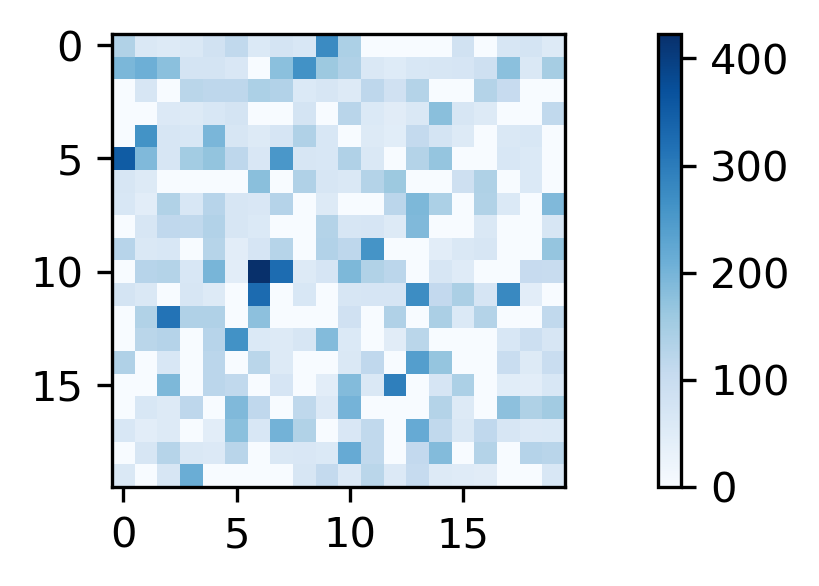}
			\caption{Category: 5}
			\label{cat_05}
	\end{subfigure}
	\hfill

	\begin{subfigure}[b]{\textwidth}
			\centering
			\includegraphics[width=0.64\linewidth]{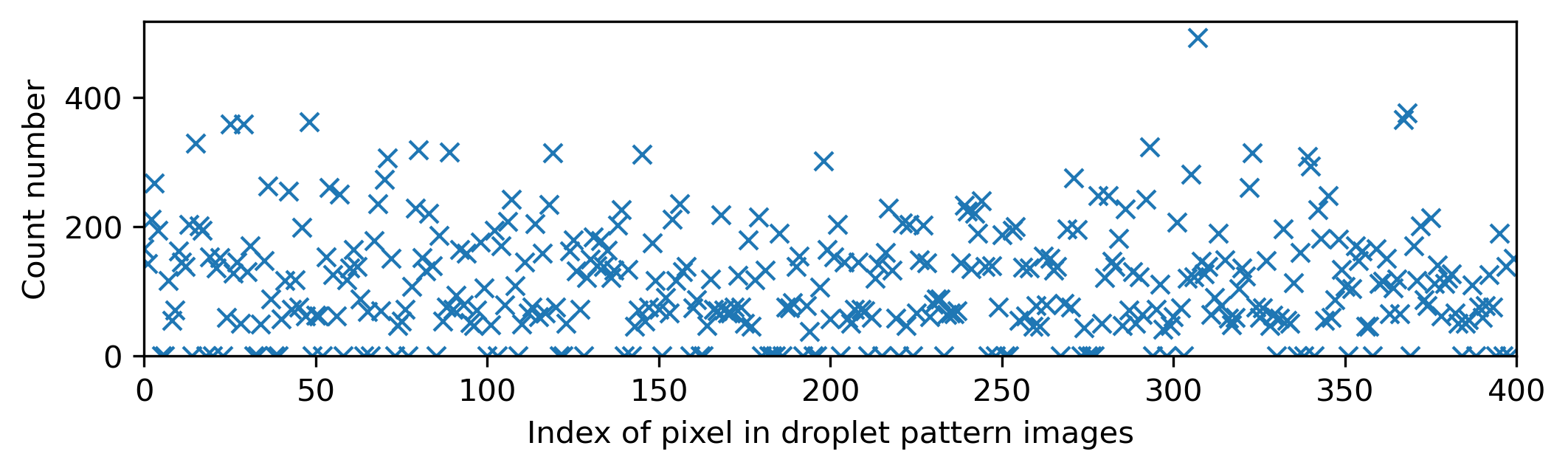}
			\includegraphics[width=0.32\linewidth]{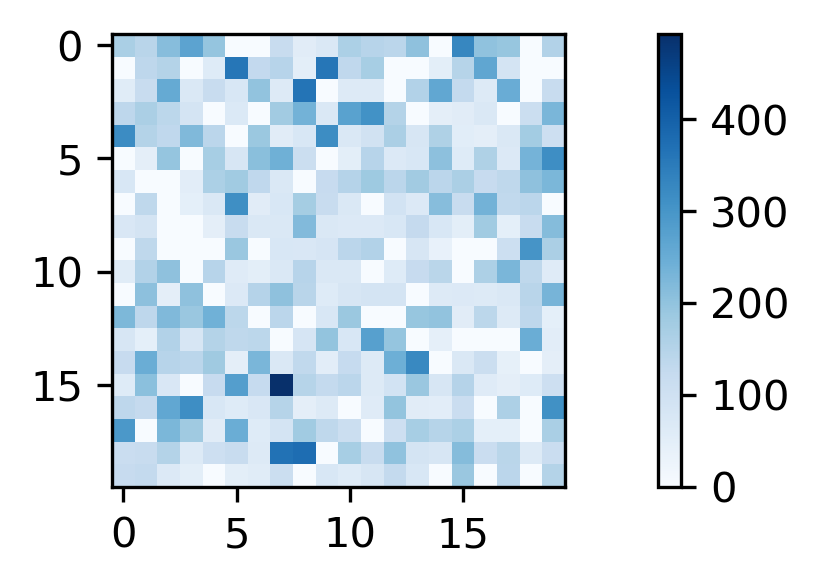}
			\caption{Category: 7}
			\label{cat_07}
	\end{subfigure}
	\hfill
	
	\begin{subfigure}[b]{\textwidth}
			\centering
			\includegraphics[width=0.64\linewidth]{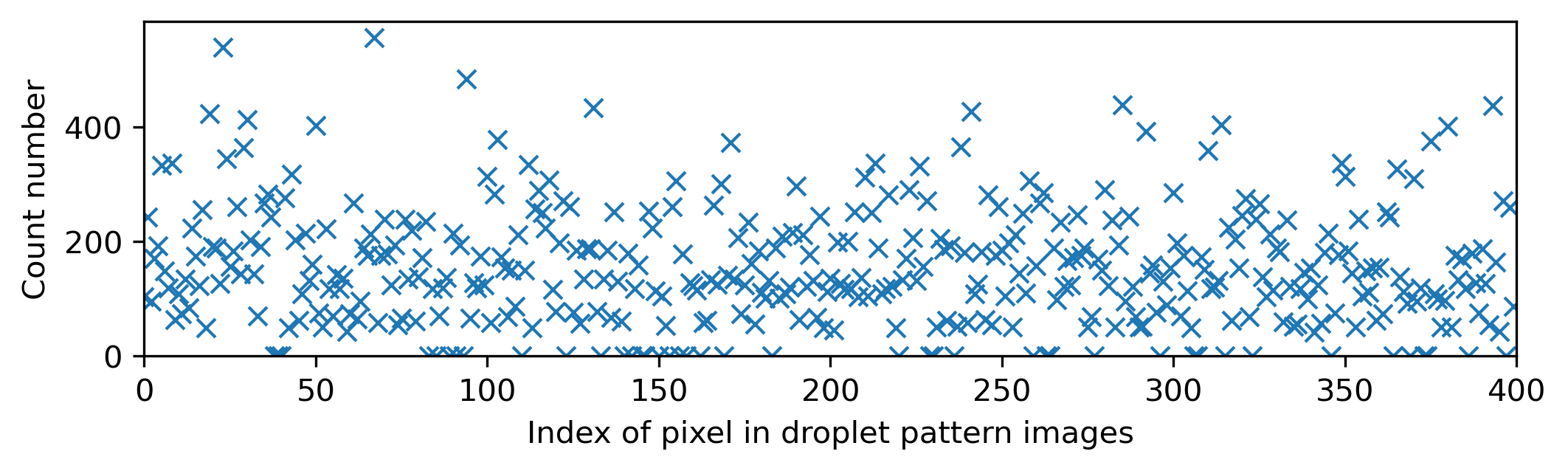}
			\includegraphics[width=0.32\linewidth]{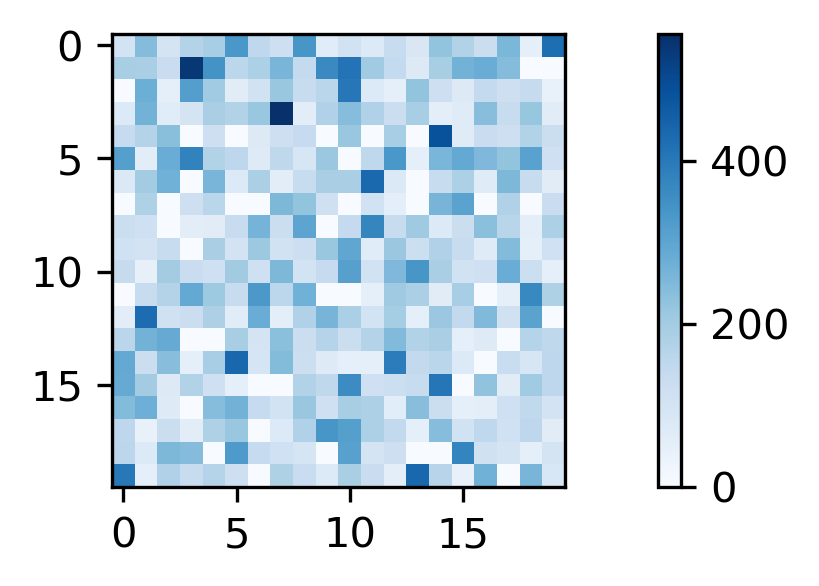}
			\caption{Category: 10}
			\label{cat_10}
	\end{subfigure}
	\hfill
	
	\begin{subfigure}[b]{\textwidth}
			\centering
			\includegraphics[width=0.64\linewidth]{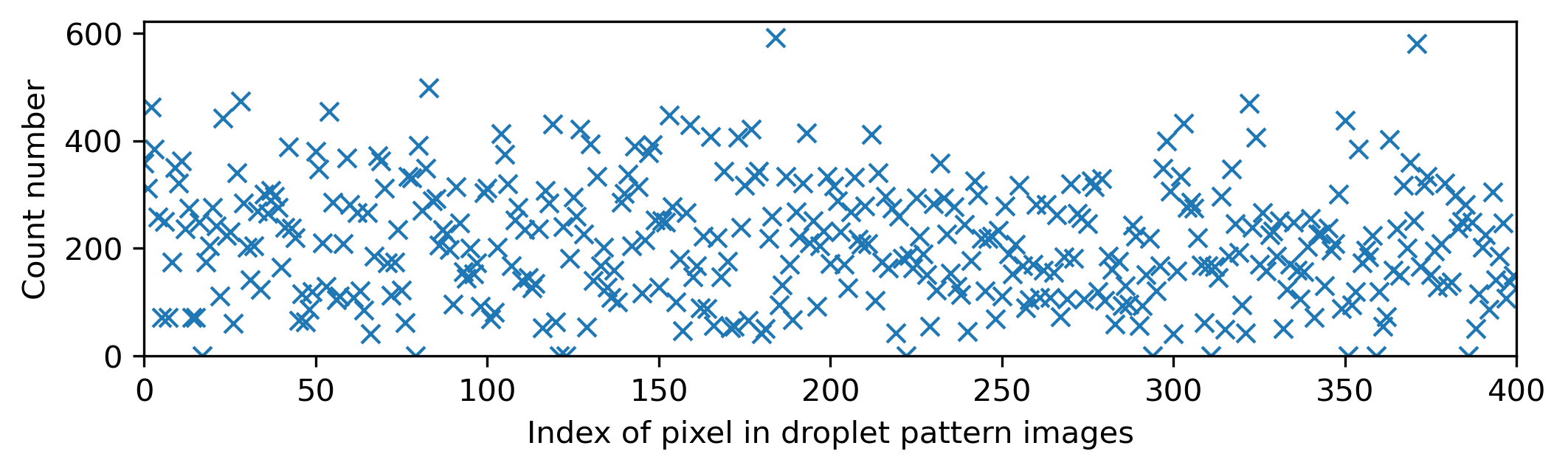}
			\includegraphics[width=0.32\linewidth]{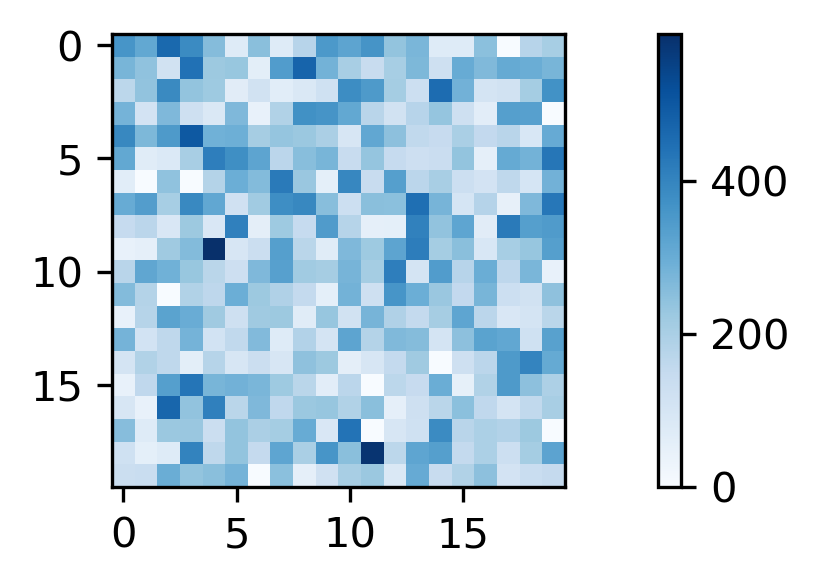}
			\caption{Category: 15}
			\label{cat_15}
	\end{subfigure}
	\hfill
			
\caption[]{(continued) 
Breakdown of the compiled datasets: the count number of occurrences for each pixel in the droplet pattern images with $20\times20$ pixels.
}
\end{figure}

\begin{figure}[!ht]
\ContinuedFloat
\centering

	\begin{subfigure}[b]{\textwidth}
			\centering
			\includegraphics[width=0.64\linewidth]{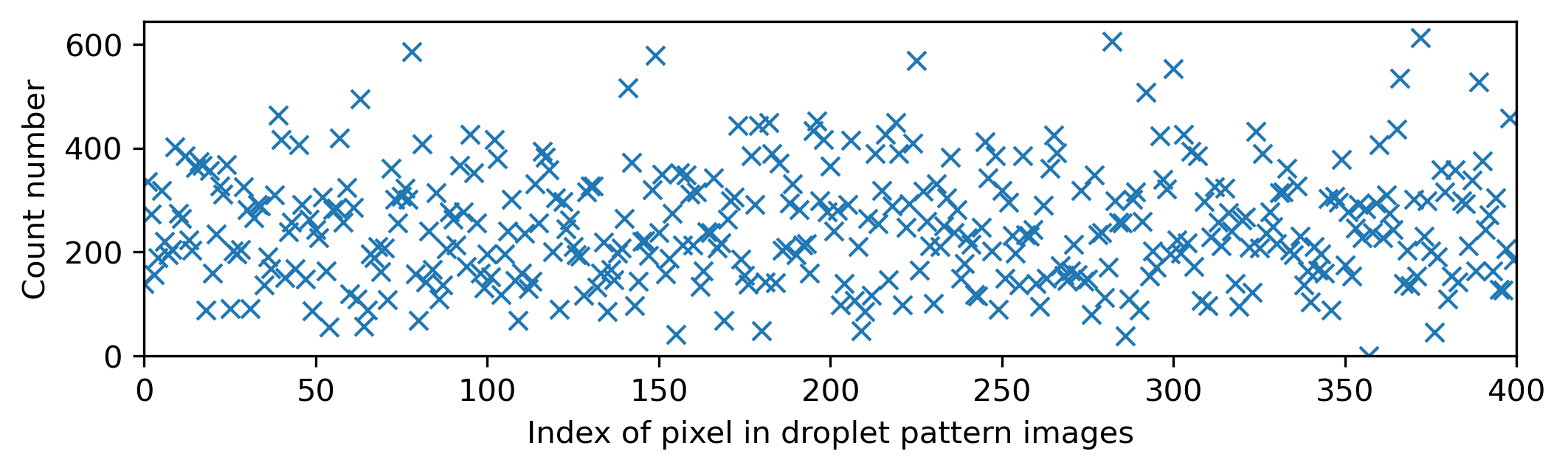}
			\includegraphics[width=0.32\linewidth]{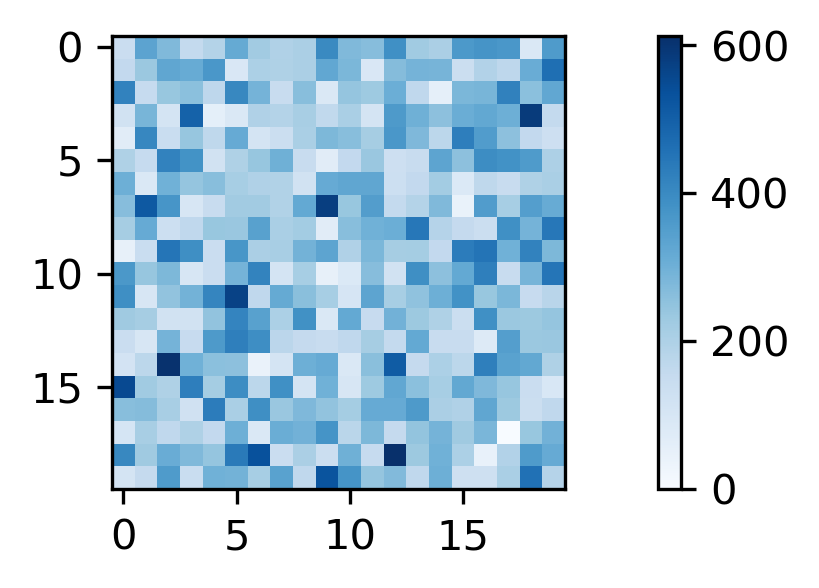}
			\caption{Category: 20}
			\label{cat_20}
	\end{subfigure}
	\hfill
	
	\begin{subfigure}[b]{\textwidth}
			\centering
			\includegraphics[width=0.64\linewidth]{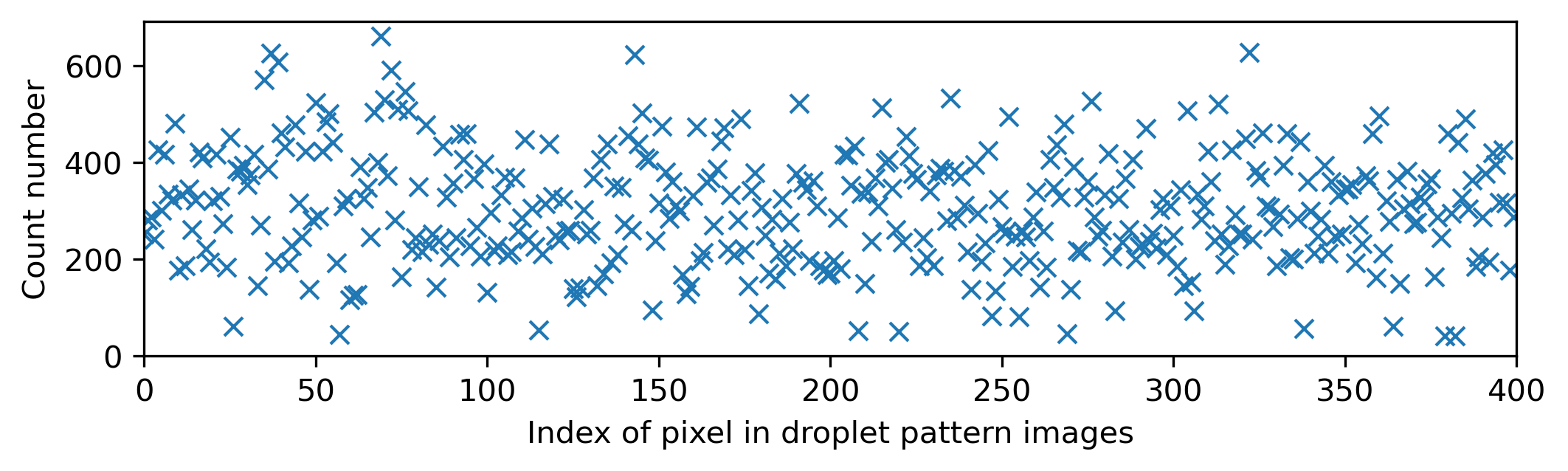}
			\includegraphics[width=0.32\linewidth]{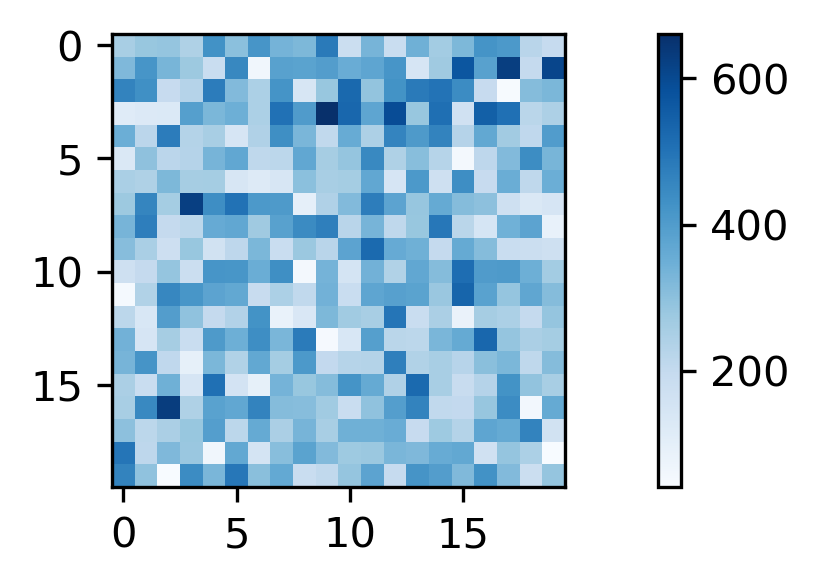}
			\caption{Category: 25}
			\label{cat_25}
	\end{subfigure}
	\hfill

	\begin{subfigure}[b]{\textwidth}
			\centering
			\includegraphics[width=0.64\linewidth]{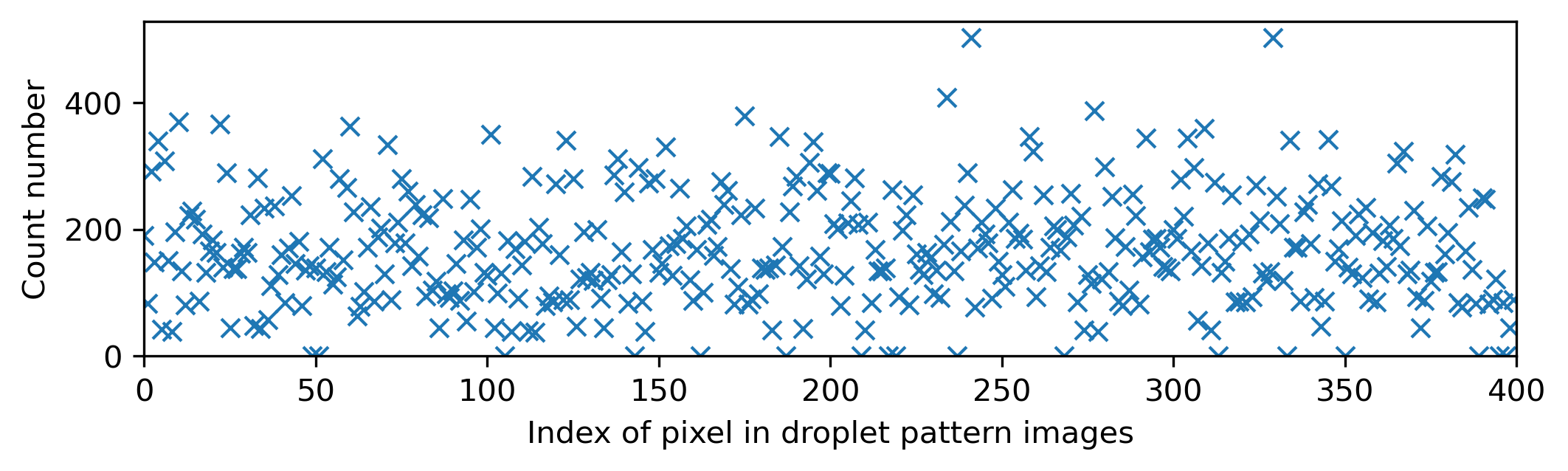}
			\includegraphics[width=0.32\linewidth]{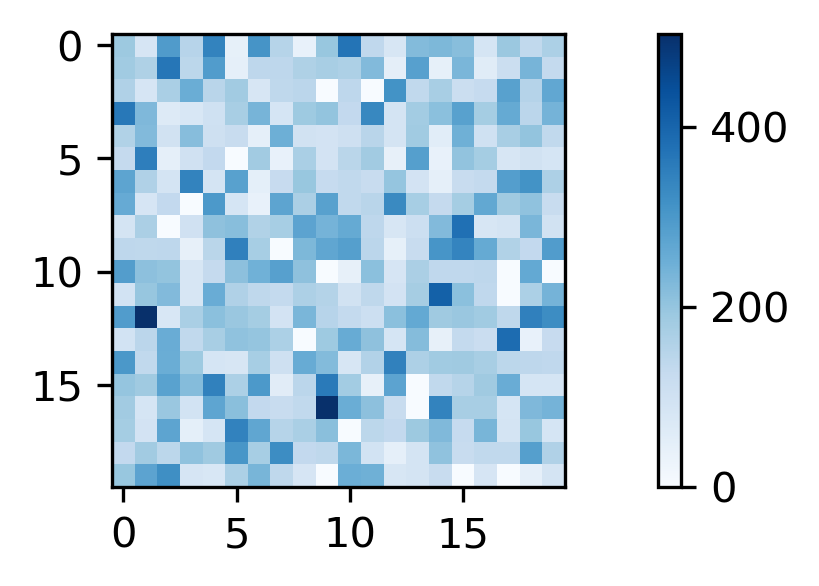}
			\caption{Category: 30}
			\label{cat_30}
	\end{subfigure}
	\hfill

	\begin{subfigure}[b]{\textwidth}
			\centering
			\includegraphics[width=0.64\linewidth]{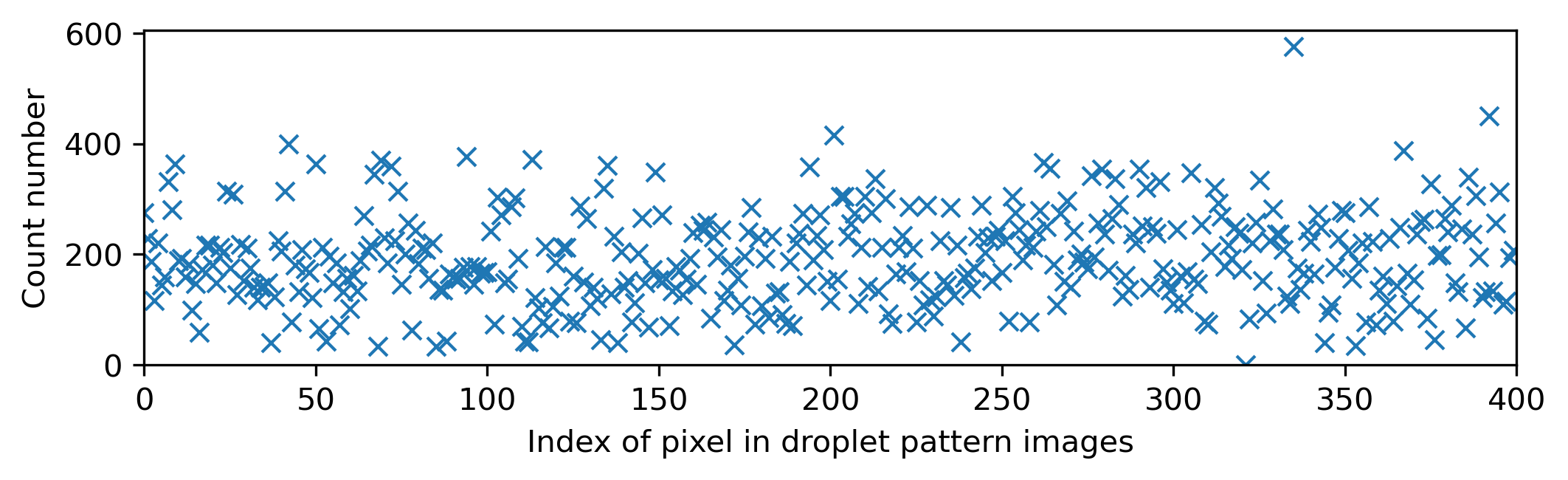}
			\includegraphics[width=0.32\linewidth]{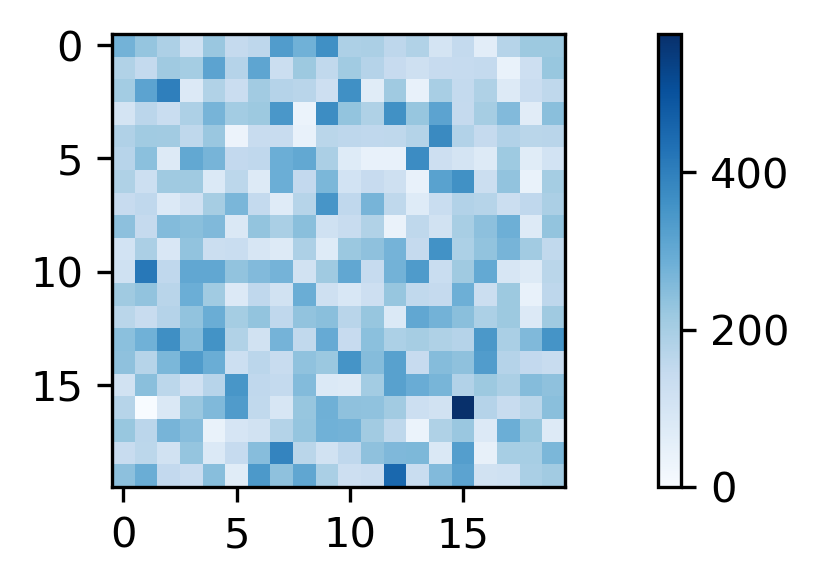}
			\caption{Category: 40}
			\label{cat_40}
	\end{subfigure}
	\hfill

\caption[]{(continued) 
Breakdown of the compiled datasets: the count number of occurrences for each pixel in the droplet pattern images with $20\times20$ pixels.
}
\end{figure}

\subsection{Structure of each dataset}
Each dataset includes four \textit{csv} files:
\begin{itemize}

\item \textbf{``t''}\textendash consisting of one column of data. The $i$\textsuperscript{th} row holds the value of spread time (unit: second) for the $i$\textsuperscript{th} example in that dataset partition.

\item \textbf{``h''}\textendash consisting of one column of data. The $i$\textsuperscript{th} row holds the value of film thickness (unit: meter) for the $i$\textsuperscript{th} example in that dataset partition.

\item \textbf{``dp''}\textendash consisting of multiple columns. The $i$\textsuperscript{th} row holds the index of all \textit{`On'} pixels (value of 1) in the droplet pattern image for the $i$\textsuperscript{th} example in that dataset partition. 
These pixels denote the firing nozzles of the printhead to dispense droplet.
Other pixels are \textit{`Off'} (value of 0).

\item \textbf{``vof''}\textendash consisting of multiple columns. The $i$\textsuperscript{th} row holds the index of all \textit{`On'} pixels (value of 1) in the imprint image for the $i$\textsuperscript{th} example in that dataset partition. 
These pixels denote the wet area of the field.
Other pixels are \textit{`Off'} (value of 0).
\end{itemize}

The developed Python package comes with some simple functions to read the dataset partition(s) from a given directory, or by walking through a root directory, and returning the compiled \textit{t}, \textit{h}, \textit{dp}, and \textit{vof} as NumPy arrays.
The \textit{dp} array has a shape of \colorbox{lightgray}{(dataset size, 400)} since each drop pattern image has 400 pixels. 
The \textit{vof} array has a shape of \colorbox{lightgray}{(dataset size, 25600)} since each imprint image has $160\times160$ pixels.

\section{A gentle dive into the fluid dynamics of the system}

\begin{figure}[b!]
\centering
\includegraphics[width=0.5\linewidth]{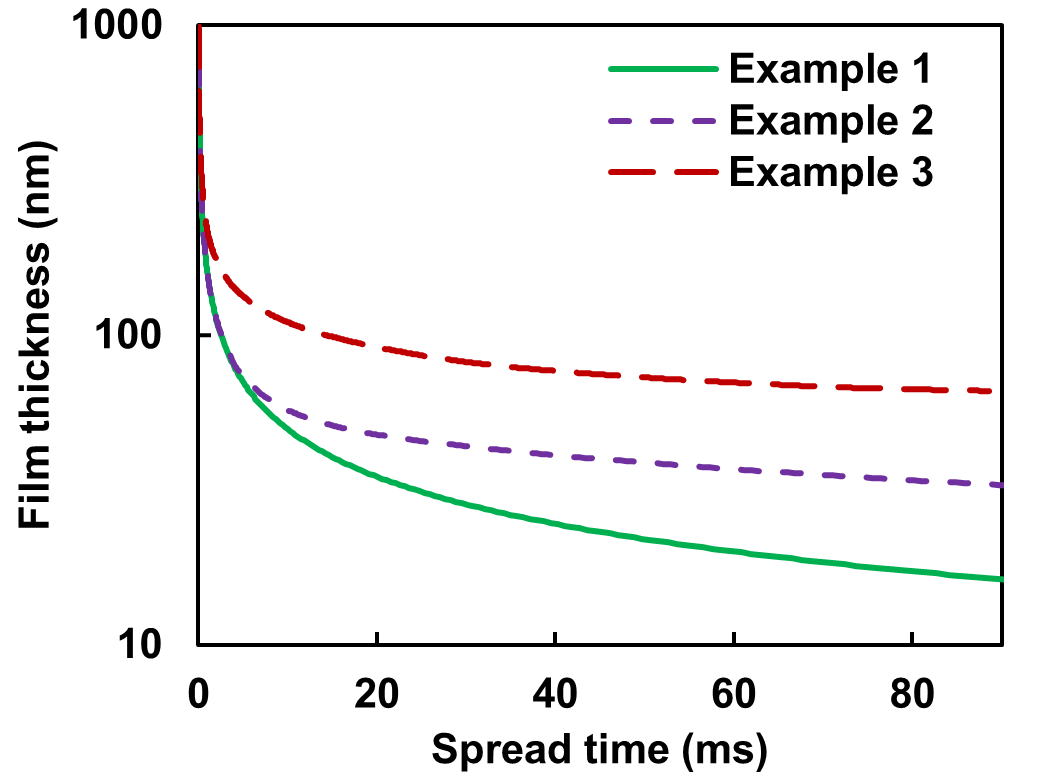}
\caption{The temporal evolution of the film thickness for three different drop pattern images presented in Fig.~\ref{fig-examples}. 
}
\label{fig-h_vs_t}
\end{figure}

\begin{figure}[tb!]
\centering
\includegraphics[width=0.7\linewidth]{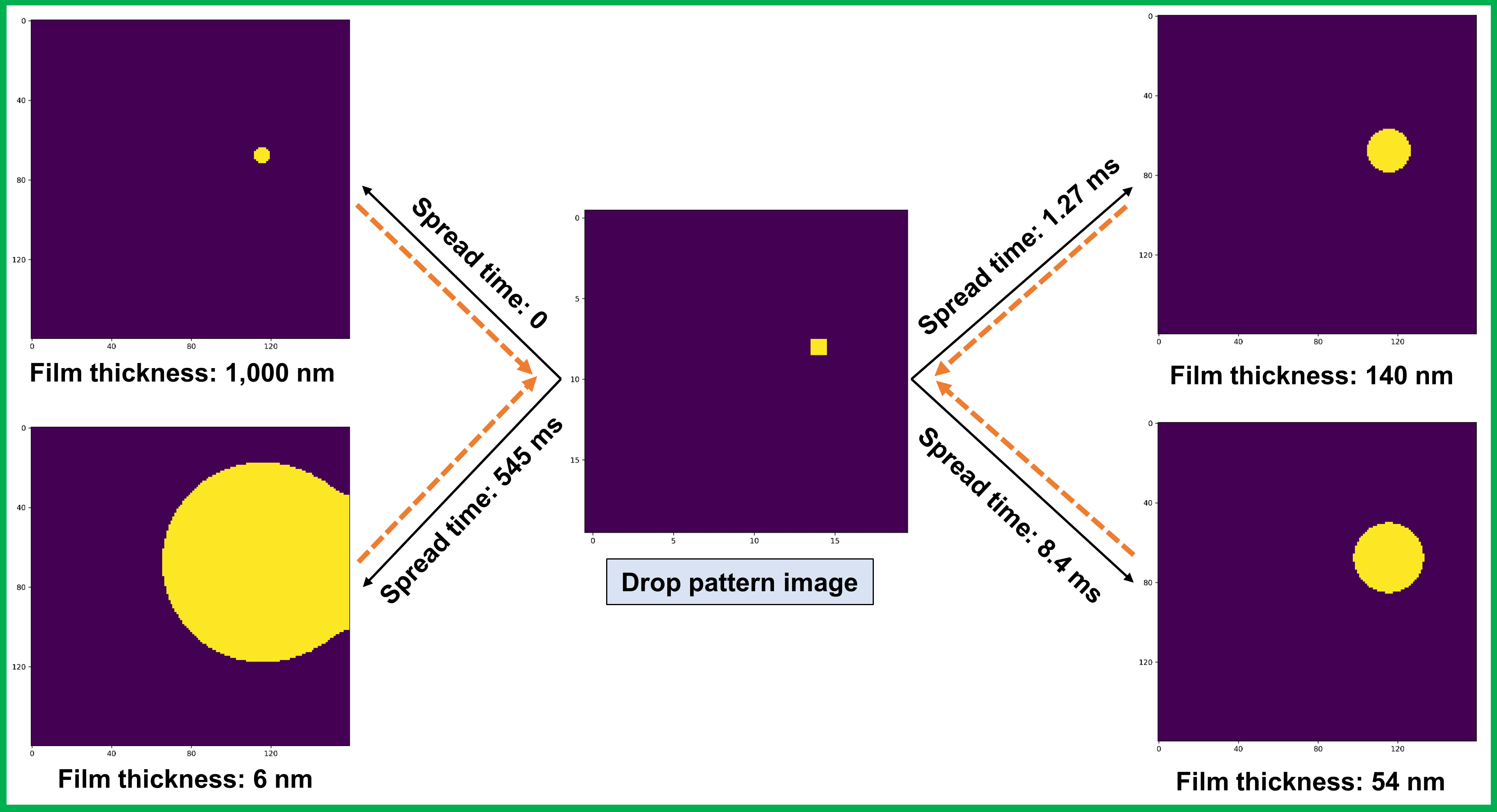}

\vspace{0.1cm}

\includegraphics[width=0.7\linewidth]{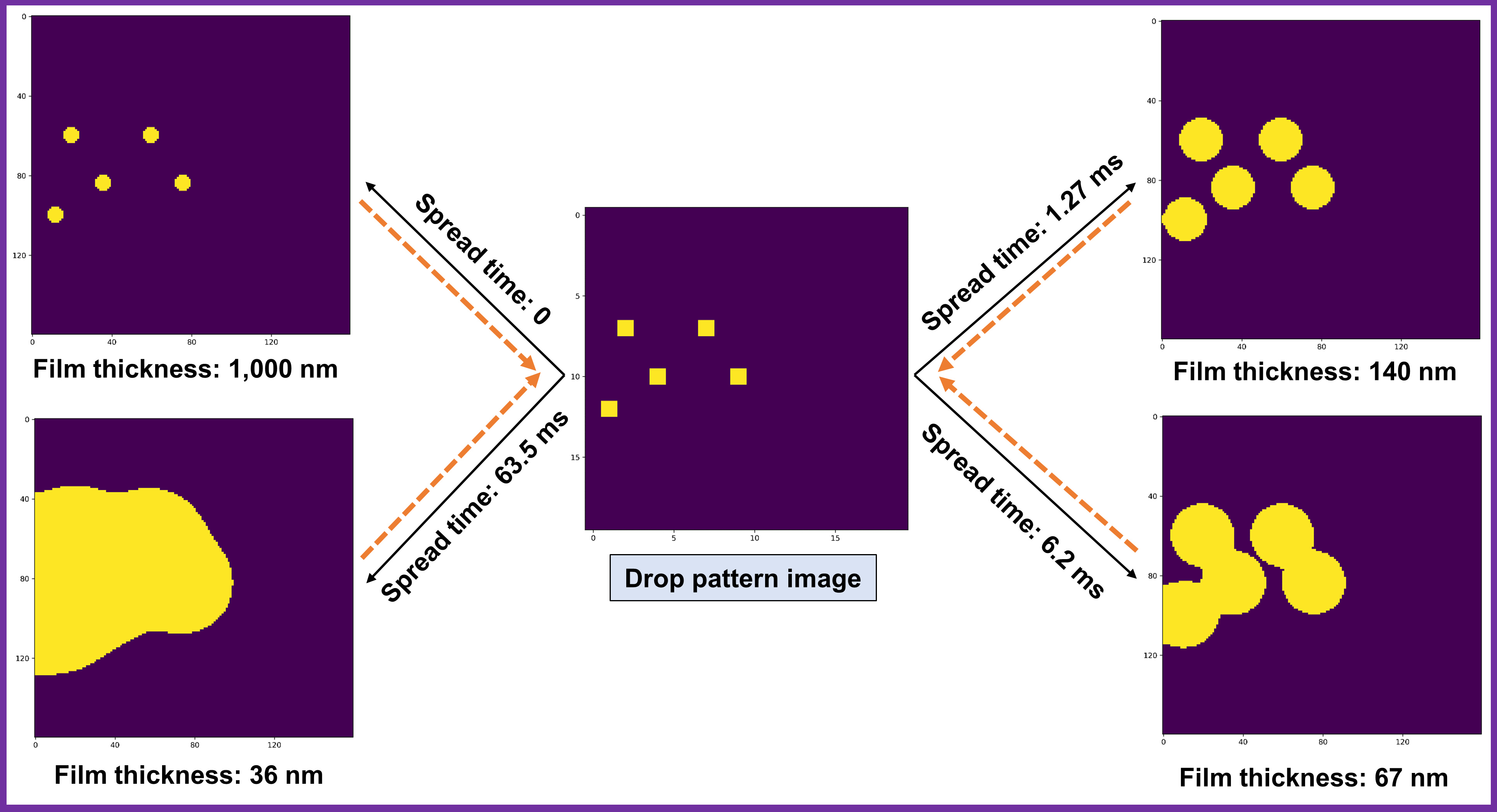}

\vspace{0.1cm}

\includegraphics[width=0.7\linewidth]{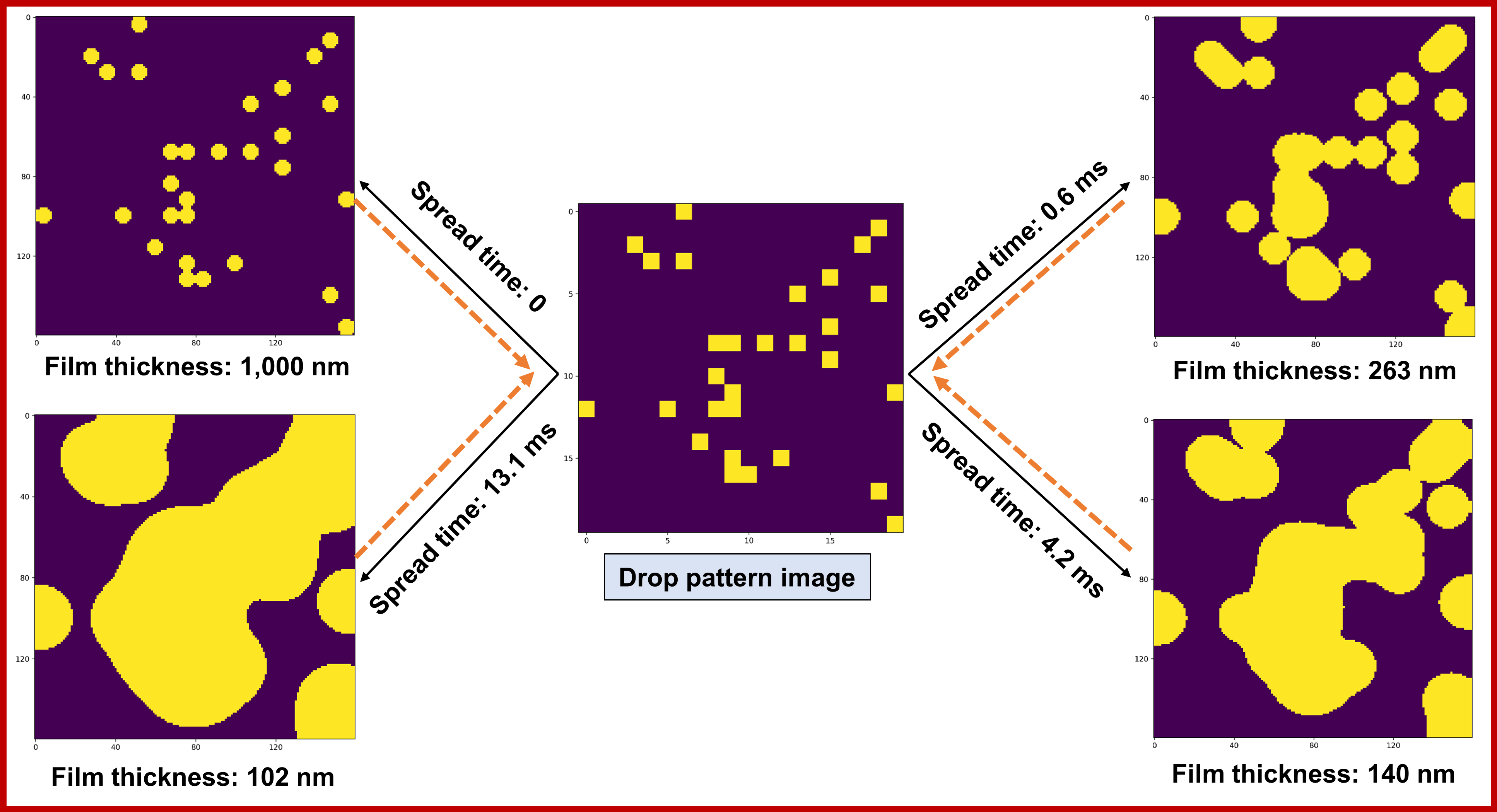}

\vspace{0.05cm}
\caption{The liquid film map (imprint image \textit{vof}) for different spread times given three different drop pattern images (\textit{dp}): \textit{Example 1} (Top), \textit{Example 2} (Middle), and \textit{Example 3} (Bottom). 
The solid black arrows show the translation of drop pattern image to an imprint image for different spread times (forward problem). 
The orange dashed arrows denote the translation of an imprint image with a given film thickness to an appropriate drop pattern image and the required spread time (inverse problem).
Here, the droplet pattern images have $20\times20$ pixels, while the imprint images contain $160\times160$ pixels.
}
\label{fig-examples}
\end{figure}

Before using the dataset, it is good to have some level of familiarity with the underlying physics of the system the dataset is derived from.
For example, an accurate and robust image-to-image mapping is not achievable by considering solely the \textit{dp} and \textit{vof} images in its entirety.
Any reasonable mapping between these images in its entirety should capture the strongly-nonlinear correlation of these images with each other and also with the spread time and liquid film thickness.
Alternatively, one may omit the effects of spread time (or liquid film thickness) by confining the dataset to a narrow range of spread time (or liquid film thickness).

In this section, we aim at explaining, from a high-level perspective, how the parameters of an example are correlated qualitatively: \textit{t} (scalar), \textit{h} (scalar), \textit{dp} (drop pattern image), and \textit{vof} (imprint image). 

Let us start with a simple system with only one droplet (\textit{Example 1}) as shown in Fig.~\ref{fig-examples} (Top). 
As we can see, only one pixel of the \textit{dp} is \textit{`On'}. 
At the beginning of the imprinting process, the imprint image resembles reasonably the drop pattern image, except that it has $160\times160$ pixels compared to $20\times20$ pixels of the drop pattern image.
As time elapses, the droplet spreads further and the liquid film thickness decreases 
(conservation of mass principle).

The dynamics of the system is relatively fast at the beginning, but it becomes slower with time.
For example, it takes about $1.27~ms$ for the liquid film thickness to reduce from $1~\mu m$ to $140~nm$, 
while it needs another $7.13~ms$, \textit{i.e.} total spread time of $8.4~ms$, to get to $54~nm$ film thickness.

Now, let us consider a \textit{dp} image with only five \textit{`On'} pixels (\textit{Example 2}) as shown in Fig.~\ref{fig-examples} (Middle).
The dynamics of the system is the same as the previous one until the droplets start to merge, \textit{i.e.} it takes about $1.27~ms$ for the liquid film thickness to reduce from $1~\mu m$ to $140~nm$.
However, the current system becomes slower than the previous one as a result of the merge of droplets.
This behavior can be discerned more easily from Fig.~\ref{fig-h_vs_t} which shows the dynamic evolution of the liquid film thickness.

Similarly, the same fluid flow characteristics can be observed for a \textit{dp} image with 29 \textit{`On'} pixels (\textit{Example 3}) as shown in Fig.~\ref{fig-examples} (Bottom), except that the merge of droplets that are, now, more in quantity and closer to each other in some areas, causes the dynamics of the system to
become slower than in \textit{Example 2}. 
Fig.~\ref{fig-h_vs_t} should help more easily observe this behavior.

In summary, it can be discerned from Figs.~\ref{fig-examples} and \ref{fig-h_vs_t} that:

\begin{itemize}
\item The dynamics of the system becomes slower with time.
\item The dynamics of the system becomes slower as a result of the merge of droplets.
\end{itemize}

\section{Network structure}

\begin{figure}[tb!]
\centering
\includegraphics[width=\linewidth]{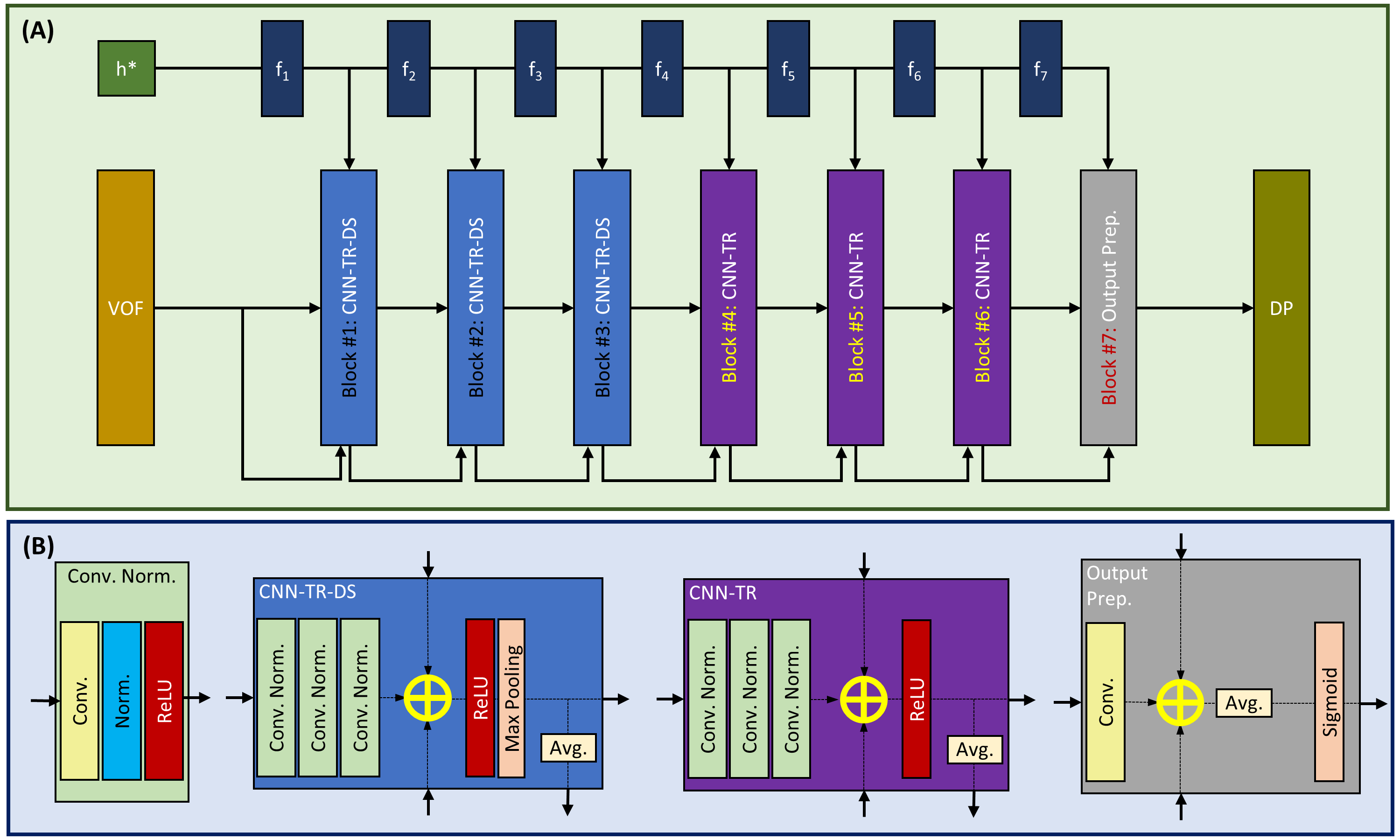}
\caption{(A) The architecture of our residual convolutional neural network with trainable and tunable refinement level (CNN-TTR) to map a high-resolution imprint image (VOF), in conjunction with a normalized film liquid thickness $h^{\ast}$, to its target low resolution droplet pattern image (DP), as well as (B) the building blocks used in our model.
}
\label{fig-network-structure}
\end{figure}

The basic principle of our proposed model is to \textbf{train} and \textbf{tune} the refinement level of convolutional layers.
At the core of this convolutional neural network with trainable and tunable refinement level (CNN-TTR) is
residual convolutional blocks with tunable refinement (CNN-TR) connected to function approximators that are trained to find the optimal refinement level.
The refinement level of CNN-TR blocks is then tuned by the trained function approximator.
A schematic representation of our proposed CNN-TTR model is depicted in Fig.~\ref{fig-network-structure}.
Our convolutional neural network consists of components to accomplish the following tasks: 
\begin{itemize}
\item \textbf{Function approximation:} dense blocks of $f_1, f_2, ...$ aiming at predicting the appropriate refinement level in each stage.

\item \textbf{Down-scaling:} a stack of convolutional layers with tunable refinement and down-scaling (CNN-TR-DS).
An input image with $160\times160$ pixels is down-scaled to 
$80\times80$ (Block \texttt{\#}1), 
$40\times40$ (Block \texttt{\#}2), 
and $20\times20$ pixels (Block \texttt{\#}3) sequentially.

\item \textbf{Widening of field of view:} a stack of convolutional layers with tunable refinement and no down-scaling (CNN-TR).

\item \textbf{Output preparation:} a convolutional layer with tunable refinement and sigmoid activation function.
\end{itemize}

\subsection{Function approximation}
The similarity between an imprint image \textit{vof} and its target drop pattern image \textit{dp} decreases as the droplets spread and merge further with time.
As a result, 
in order to predict the target \textit{dp} image, 
more refinement (removing `On' pixels properly) of the \textit{vof} image is required.
Finding the appropriate refinement level in each stage is accomplished by a function approximator. 
Such function approximators aim at finding the appropriate refinement levels to be used by their corresponding residual convolutional blocks with tunable refinement (CNN-TR and CNN-TR-DS).
Here, a function approximator \cite{hornik_multilayer_1989} is a dense neural network with one hidden layer of 128 units and activation function of ReLU \cite{glorot_deep_2011}.

\subsection{Convolutional blocks with tunable refinement with (CNN-TR-DS) and without (CNN-TR) down-scaling}
Each CNN-TR and CNN-TR-DS block has three inputs: 
1. feature maps from the preceding block,
2. a single map consisting of the element-wise average of the feature maps from the preceding block,
and 3. a refinement level indicator provided by the corresponding function approximator.
Providing the average of the feature maps of the last preceding block enables access to the corresponding hierarchical features and mitigate the loss of information that should be preserved. 

Each CNN-TR and CNN-TR-DS block consists of a stack of three connected \textit{Conv. Norm.} blocks.
The output of this stack is added (element-wise) to inputs \texttt{\#}2 and 3: 
the refinement level indicator and the average feature map from the last preceding block.
The difference of CNN-TR and CNN-TR-DS is that the latter uses a max pooling layer for down-scaling, 
while CNN-TR blocks are used to widen the field of view.

Each \textit{Conv. Norm.} block comprises a convolutional layer with the kernel size of 5, 
followed by an instance normalization \cite{ulyanov_instance_2017}, and ReLU activation function \cite{glorot_deep_2011}.
For the CNN-TR-DS blocks, each convolutional layer has 32, 64, and 128 filters in Blocks \texttt{\#}1, 2, and 3, respectively.
For the CNN-TR blocks, each convolutional layer has 128, 64, and 32 filters in Blocks \texttt{\#}4, 5, and 6, respectively.

\subsection{Output preparation block}
Block \texttt{\#}7 is similar to a CNN-TR block with the difference that it has only one convolutional layer with no normalization or ReLU activation.
It has a sigmoid activation function at the end of the block.
The convolutional layer in this block has 8 filters with a kernel size of 5.

\section{Results and discussion}
\label{results}

\subsection{Pre-processing}
\label{prep-processing}
When preparing the training, validation, and test datasets, we selected data from different categories according to Table~\ref{tab_train_val_test_datasets}.
\bgroup
\def\arraystretch{1.5}
\begin{table}[b!]
\centering
\begin{tabular}{|| lcccccccccccc ||}
\hline
Category & 1 & 2 & 3 & 4 & 5 & 7 & 10 & 15 & 20 & 25 & 30 & 40 \\ [0.5ex] 
\hline
\hline
Training & 0.125 & - & - & 1.0 & - & 1.0 & 1.0 & - & 1.0 & 1.0 & - & 1.0 \\[0.5ex] 
Validation & 0.125 & - & - & - & 1.0 & - & - & 1.0 & - & - & - & - \\[0.5ex] 
Test & 0.125 & - & - & - & - & - & - & - & - & - & 1.0 & - \\[0.5ex] 
\hline
\end{tabular}
\caption{Fraction of each category selected for pre-processing when preparing training, validation, and test datasets.}
\label{tab_train_val_test_datasets}
\end{table}
\egroup
To mitigate the issue with examples that can potentially have multiple solutions, 
we omitted the examples with too large continuous liquid film in their \textit{vof} image. 
We displaced an interrogation window with $72\times72$ pixels across each \textit{vof} image.
Any example with a local coverage greater than $90\%$ was removed from the datasets. 
The final number of examples in training, validation, and test datasets are shown in Table.~\ref{tab_train_val_test_datasets_after_preprocessing}.
\bgroup
\def\arraystretch{1.5}
\begin{table}[b!]
\centering
\begin{tabular}{|| ccc||}
\hline
Training & Validation & Test \\ [0.5ex] 
\hline
\hline
12,476 & 5,703 & 2193 \\[0.5ex] 
\hline
\end{tabular}
\caption{Number of examples in training, validation, and test datasets after pre-processing.}
\label{tab_train_val_test_datasets_after_preprocessing}
\end{table}
\egroup

\begin{figure}
\centering
	\begin{subfigure}[b]{0.3\textwidth}
			\centering
			\includegraphics[width=\textwidth]{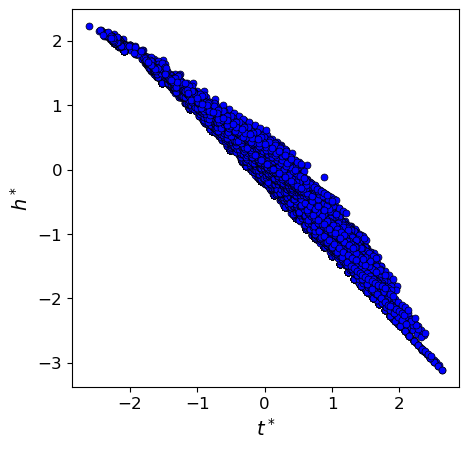}
			\caption{Training dataset}
			\label{train}
	\end{subfigure}
	\hfill
	\begin{subfigure}[b]{0.3\textwidth}
			\centering
			\includegraphics[width=\textwidth]{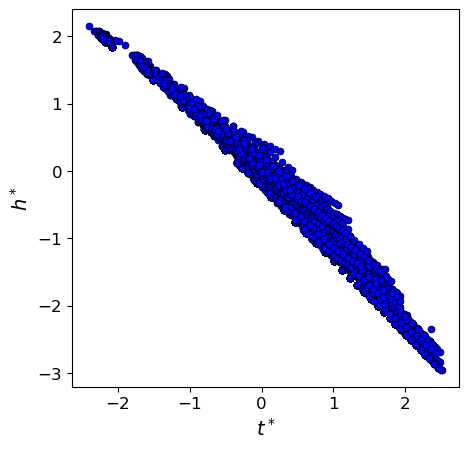}
			\caption{Validation dataset}
			\label{validation}
	\end{subfigure}
	\hfill
	\begin{subfigure}[b]{0.3\textwidth}
			\centering
			\includegraphics[width=\textwidth]{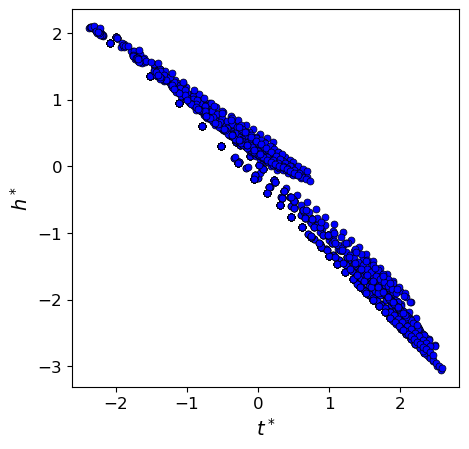}
			\caption{Test dataset}
			\label{test}
	\end{subfigure}
\caption{
The distribution of $h^{\ast}$ and $t^{\ast}$ for the training, validation, and test datasets.
}
\label{fig-dataset_h_vs_t}
\end{figure}

We use $t^{\ast}$ and $h^{\ast}$ as the normalized scalars for $t$ and $h$, respectively, as defined below:

\begin{equation}
\begin{split}
t^{\ast}&=\frac{\ln{t}-\mu_t}{\sigma_t}\\[5pt]
h^{\ast}&=\frac{\ln{h}-\mu_h}{\sigma_h},
\end{split}
\end{equation}
wherein $\mu_t$ and $\sigma_t$ refer to the arithmetic mean and standard deviation of $\ln{t}$ for the examples within the training dataset, respectively.
Similarly, $\mu_h$ and $\sigma_h$ refer to the arithmetic mean and standard deviation of $\ln{h}$ for the examples within the training dataset, respectively.
The distribution of $h^{\ast}$ and $t^{\ast}$ is shown in Fig.~\ref{fig-dataset_h_vs_t} for the training, validation, and test datasets.

\subsection{Training}
\label{training}
The model has 3,085,935 trainable parameters in total.
We used TensorFlow \cite{abadi_tensorflow_2016} and Keras \cite{chollet_keras_2015} to train our model. 
We considered the binary cross entropy loss while applying a $L_2$ regularization penalty to bias and kernel of the convolutional layers.

We use the precision-recall (PR) plot and the associated area under curve (AUC) metric to evaluate the performance of our model.
In average, only about $3\%$ of pixels of \textit{dp} images are `On'.
We noticed that  in our case, because of having an imbalanced dataset, the AUC of receiver operating characteristics (ROC) misleadingly leads to a much higher value than the AUC-PR.
Therefore, the AUC-ROC could be deceptive when drawing conclusions about the performance of the model on 
the current dataset wherein the number of negatives outweighs the number of positives significantly \cite{saito_precision-recall_2015}.

\begin{figure}
\centering
	\begin{subfigure}[b]{0.49\textwidth}
			\centering
			\includegraphics[width=\textwidth]{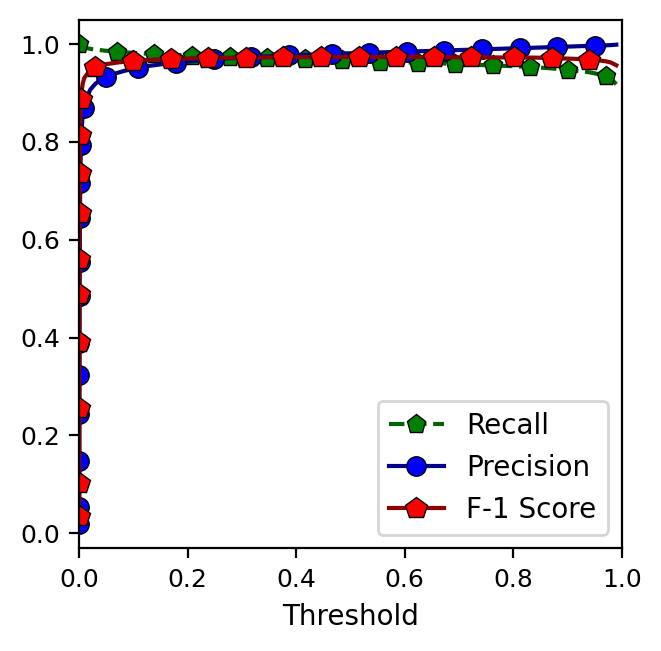}
			\caption{}
			\label{thresholds}
	\end{subfigure}
	\hfill
	\begin{subfigure}[b]{0.49\textwidth}
			\centering
			\includegraphics[width=\textwidth]{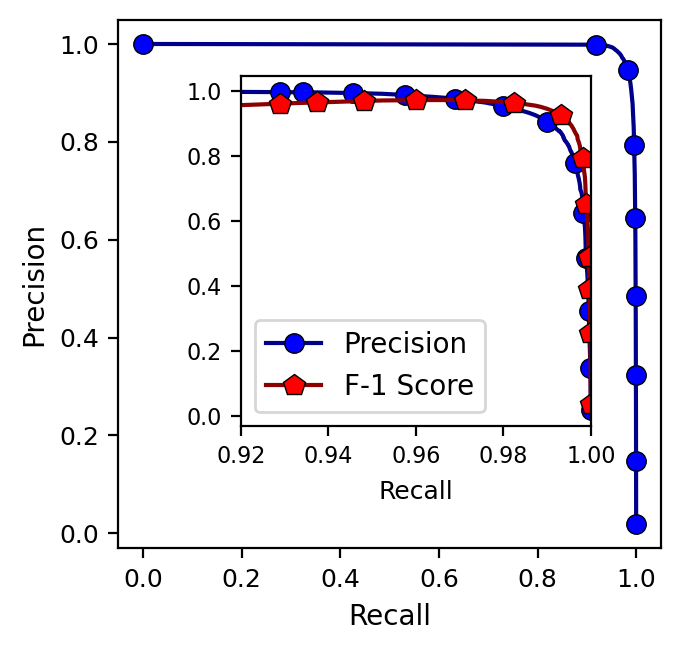}
			\caption{}
			\label{PR}
	\end{subfigure}

\caption{
(a) The variations of precision, recall, and F-1 score with the classification threshold when running the model on the validation dataset.
(b) The precision-recall plot as well as (Inset) the F-1 score variations for the magnified region of interest.
}
\label{fig-AUC_PR_F1}
\end{figure}

The validation dataset was used to examine the performance of the trained model for various classification thresholds.
The results are shown in Fig.~\ref{fig-AUC_PR_F1} (a) in terms of precision, recall, and F-1 Score.
The PR curve is also presented in Fig.~\ref{fig-AUC_PR_F1} (b).
The obtained AUC-PR for the validation dataset equals $0.9956$.
The PR curve shows an acceptable performance over a wide range of thresholds, but the highest F-1 score related to the validation dataset is found when the threshold is about $0.505$. 
The corresponding F-1 score is $0.9735$.
Finally, by using this optimized model and classification threshold, we obtain the F-1 score of $0.9647$ on the test dataset.

We used the test dataset to compare the trained model's performance with that of a crude model consisting of three max pooling layers with no trainable parameters.
The results are shown in Table.~\ref{tab_baseline_vs_model_test_dataset}.
\bgroup
\def\arraystretch{1.5}
\begin{table}[b!]
\centering
\begin{tabular}{|| lcccc ||}
\hline
Model & AUC-PR & Precision & Recall & F-1 \\ [0.5ex] 
\hline
\hline
Baseline crude model & 0.1159 & 0.1159 & 1.0000 & 0.0886 \\[0.5ex] 
CNN-TTR & 0.9951 & 0.9734 & 0.9562 & 0.9647 \\[0.5ex] 
\hline
\end{tabular}
\caption{Comparison of the performances of the model CNN-TTR and a crude model consisting of three max pooling layers without any trainable parameter when acting on the test dataset.
Here, the threshold equals 0.5 when calculating the precision, recall, and F-1 score.
}
\label{tab_baseline_vs_model_test_dataset}
\end{table}
\egroup
As expected, the crude model can predict all the `ON' pixels of the \textit{dp} images (Recall=1.0) at the cost of a relatively low precision of about 11.59\%.

\subsection{Function approximators}
\begin{figure}
\centering
	\begin{subfigure}[b]{0.49\textwidth}
			\centering
			\includegraphics[width=\textwidth]{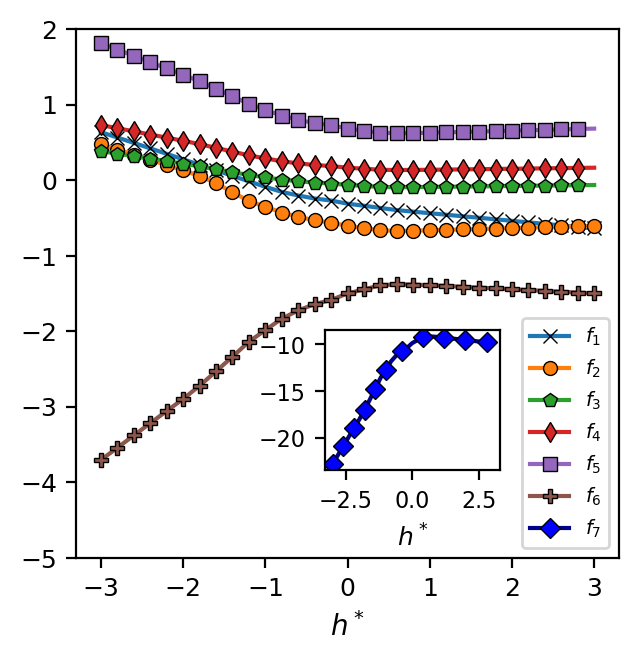}
			\caption{}
			\label{func_approx_ref}
	\end{subfigure}
	\hfill
	\begin{subfigure}[b]{0.49\textwidth}
			\centering
			\includegraphics[width=\textwidth]{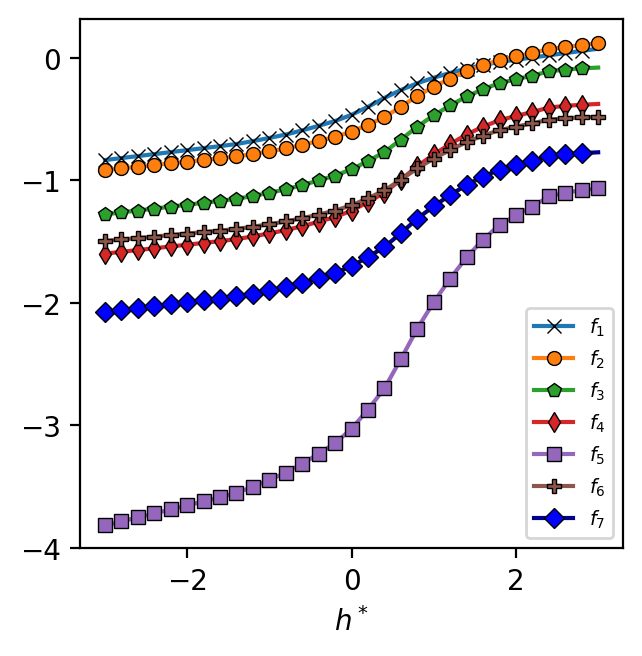}
			\caption{}
			\label{func_approx_linear}
	\end{subfigure}

\caption{
(a) The output of function approximators for various $h^{\ast}$ values using the trained model CNN-TTR and (b) its counterpart 
with a linear activation function within the CNN-TR and CNN-TR-DS blocks and 
no normalization layer in the \textit{Conv. Norm.} blocks.
}
\label{fig-func_approx}
\end{figure}

The output of function approximators $f_1, f_2, ..., f_7$ is shown in Fig.~\ref{fig-func_approx} for various $h^{\ast}$ values.
Fig.~\ref{fig-func_approx} (a) shows the results for the trained model CNN-TTR the schematic representation of which was shown in Fig.~\ref{fig-network-structure}. 
When drawing conclusions about these results, we need to keep in mind that the model CNN-TTR consists of an instance normalization layer within the \textit{Conv. Norm.} blocks and the nonlinear activation function of ReLU within the CNN-TR and CNN-TR-DS blocks, as a result of which the observed trends my not be easily interpretable.
Nevertheless, the plot shows that, in general, the outputs are negative. 
Together with the immediate ReLU activation function within the CNN-TR and CNN-TR-DS blocks, such negative translations lead to a sequential refinement, 
\textit{i.e} removal of `On' pixels of image in a sequential fashion.

Fig.~\ref{fig-func_approx} (b) shows the results for a less effective (slower training and lower AUC-PR) variant of CNN-TTR with a linear activation function within the CNN-TR and CNN-TR-DS blocks and 
no normalization layer within the \textit{Conv. Norm.} blocks.
Because of omitting the non-linearity associated with the aforementioned components,
the behavior of function approximators is, now, more interpretable:
a \textit{vof} image with a thinner liquid film (smaller $h^{\ast}$) requires more refinement to obtain its target \textit{dp} image.
Hence, the function approximators output a smaller value (negative number with a larger magnitude) as $h^{\ast}$ decreases to lead to a larger refinement level (removal of more `On' pixels).

\subsection{Sequential refinement}
\label{seq-refinement}
\begin{figure}
\centering
\includegraphics[width=\textwidth]{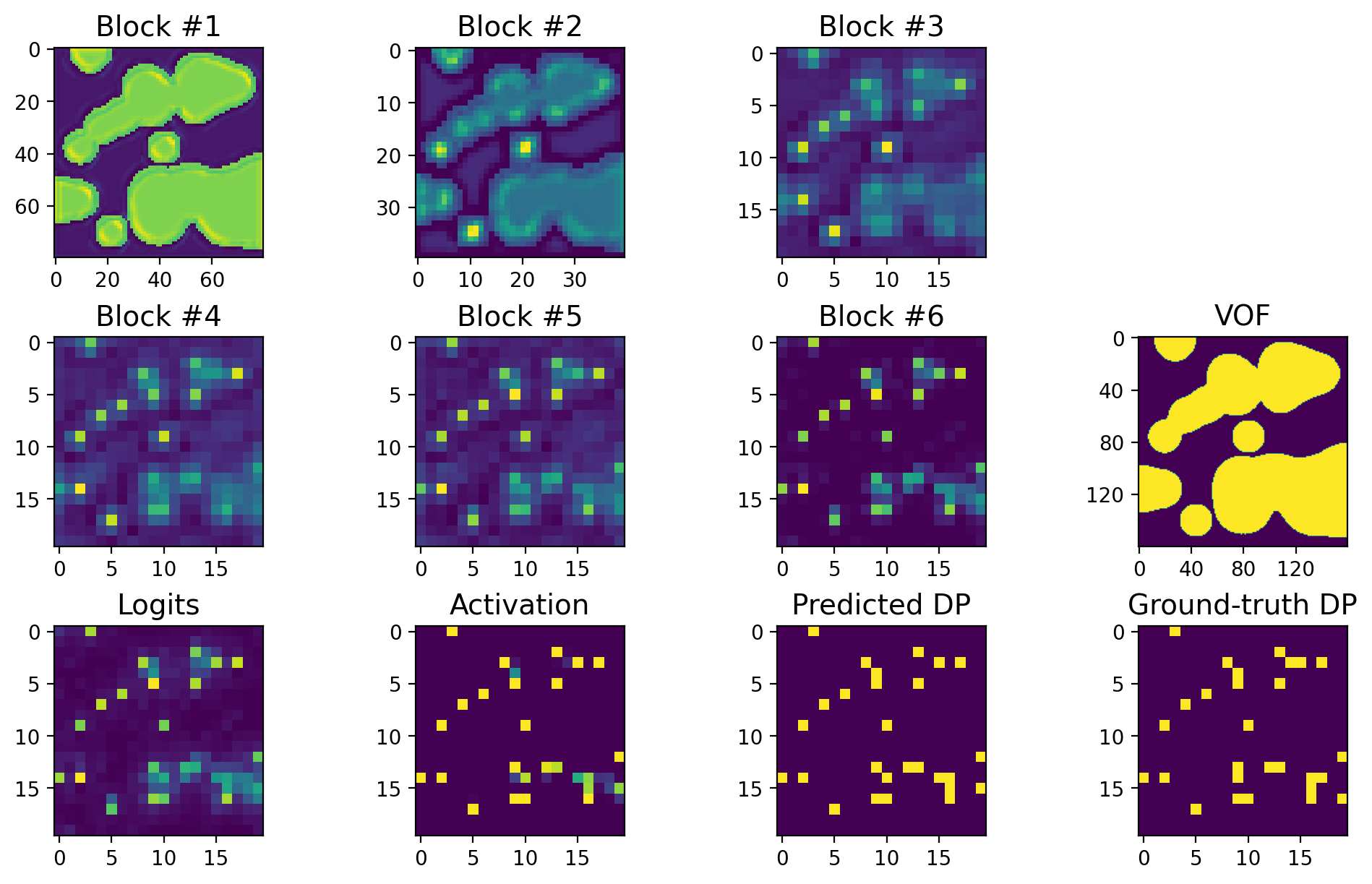}
\caption{
The average response maps of Blocks \texttt{\#}1\textendash6 as well as the logits and the sigmoid activation map related to Block \texttt{\#} 7, together with the ground-truth \textit{dp} image for a given example from the test dataset with $h^{\ast}=-0.001061$.
}
\label{fig-seq_refinement_ex_5483}
\end{figure}

The concept of sequential refinement is demonstrated in Fig.~\ref{fig-seq_refinement_ex_5483} for a given example from the test dataset.
The model refines the \textit{vof} image in multiple steps based on the normalized thickness of the liquid film $h^{\ast}$ until the predicted \textit{dp} image is obtained.
It can be discerned that a region with a relatively small puddle of liquid is more easily refined, 
while an area with a large continuous liquid film is more challenging to be refined properly.

There are two potential culprits for the problem associated with the large-coverage regions:
1. insufficient examples during the training of model, and
2. non-uniqueness of solution for such regions.
In particular, a large-coverage region my be obtainable by multiple droplet patterns that differ only slightly in one or a few pixels (usually close to the center of such large-coverage regions).

\begin{figure}[tb!]
\centering
\includegraphics[width=0.8\textwidth]{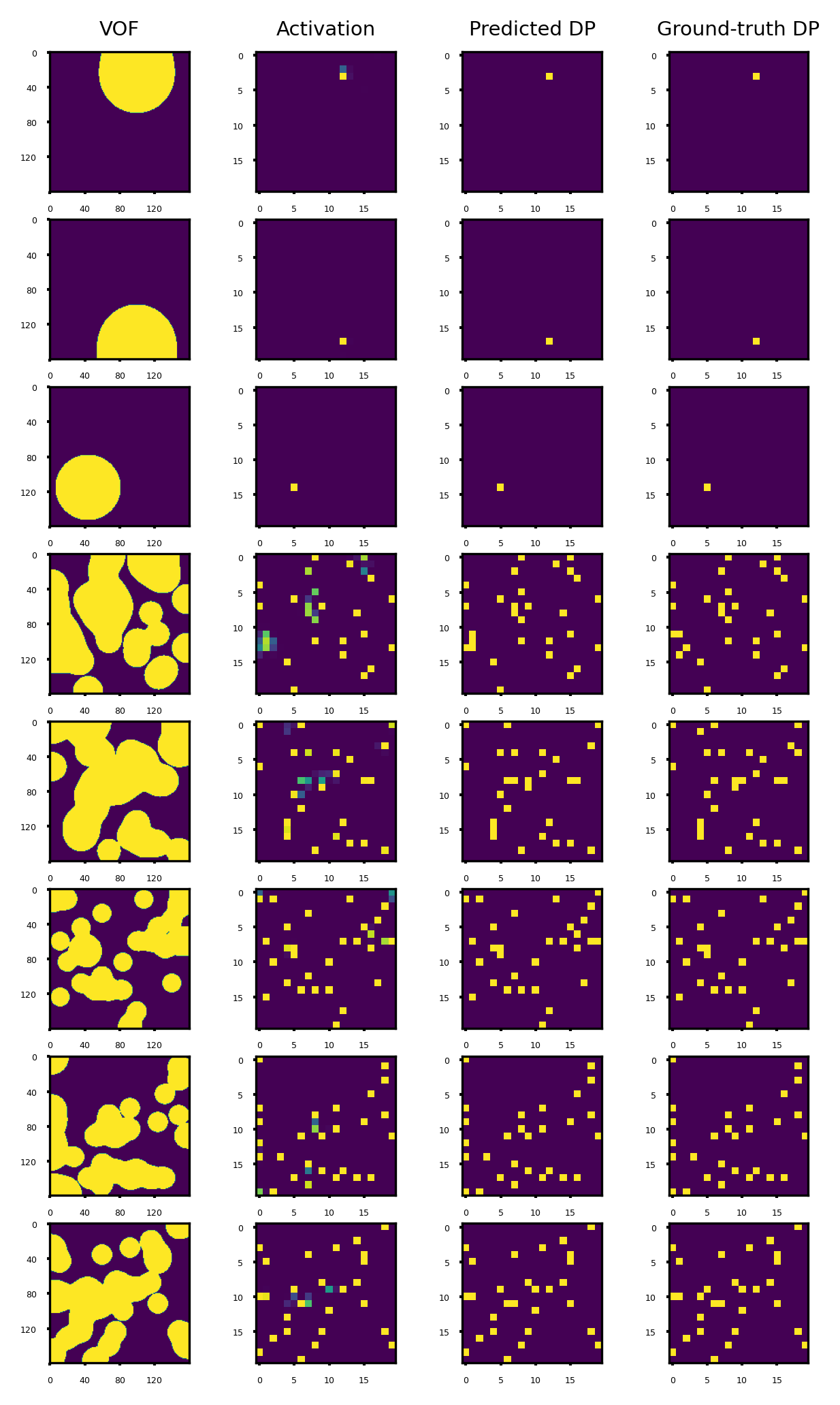}
\caption{
Prediction of the model CNN-TTR for several examples from the test dataset.
Here, the classification threshold equals 0.4 when obtaining the ``Predicted DP'' from the sigmoid layer's activation.
From top to bottom, $h^{\ast}$ is -2.641, -2.574, -2.435, -0.173, -0.136, 0.317, 0.169, and 0.123.
}
\label{fig-seq_refinement_ex_test_list}
\end{figure}

One of the consequences of the model getting less confident in these situations is that some of the `On' pixels can be lost when using the classification threshold optimized for the highest F-1 score. 
In such situations, reducing the classification threshold can lead to a higher recall value at the cost of reducing the precision and F-1 score.
However, as shown in Fig.~\ref{fig-AUC_PR_F1}, the model performs acceptably well on a wide range of the classification threshold.
Hence, one can reduce the classification threshold to improve the recall metric.
We have noticed that the subsequent reduction of precision is partly due to the slight misplacement of one or few `On' pixels close to the center of large-coverage regions, 
which is still much better than completely loosing those `On' pixels.
The results obtained using a classification threshold of 0.4 are shown in Fig.~\ref{fig-seq_refinement_ex_test_list}
for some of the examples from the test dataset.

\subsection{Kernels}
\label{kernels}

Some of the kernels of the first convolutional layer in Blocks \texttt{\#}1\textendash7 are shown in Fig.~\ref{fig-kernels}.
\begin{figure}[!ht]
\captionsetup{belowskip=10pt}
\centering
\subfloat[Block \texttt{\#}1]{\includegraphics[width=0.8\textwidth]{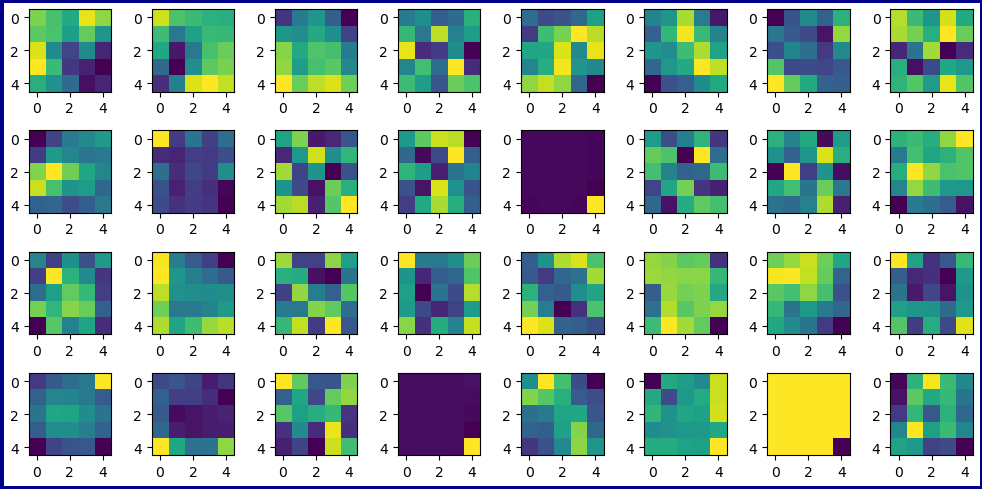}}
\hfill
\subfloat[Block \texttt{\#}2]{\includegraphics[width=0.8\textwidth]{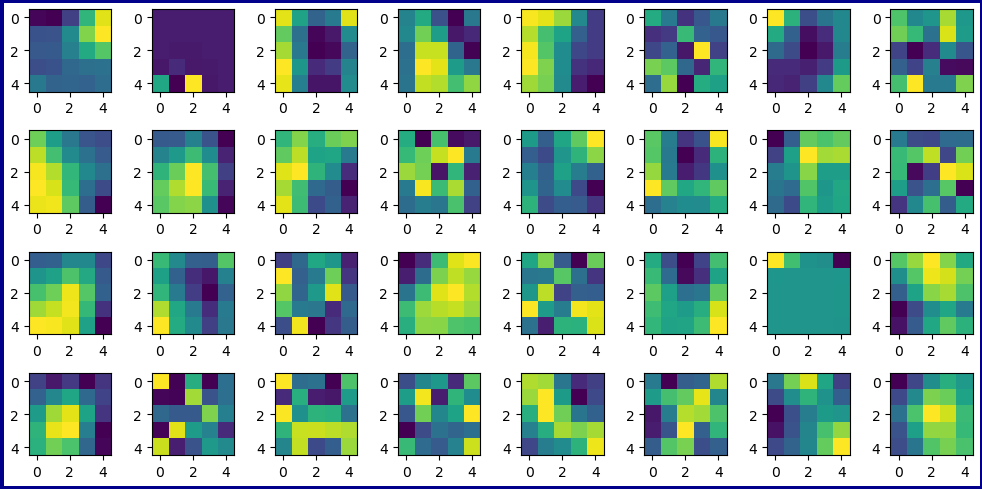}}
\hfill
\subfloat[Block \texttt{\#}3]{\includegraphics[width=0.8\textwidth]{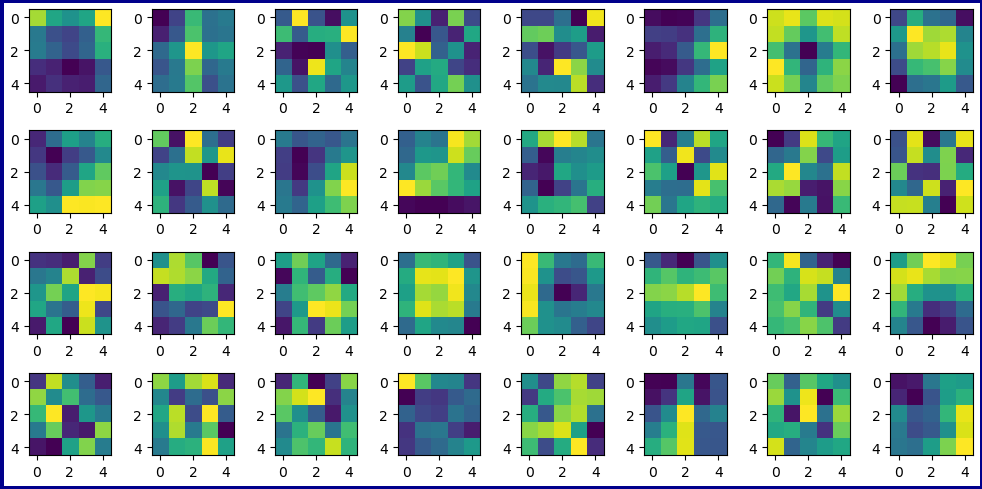}}
	
\caption{
Some of the kernels of the first convolutional layer for different Blocks (normalized to vary between 0 and 1).
}
\label{fig-kernels}
\end{figure}
\begin{figure}[!ht]
\captionsetup{belowskip=10pt}
\ContinuedFloat
\centering
\subfloat[Block \texttt{\#}4]{\includegraphics[width=0.8\textwidth]{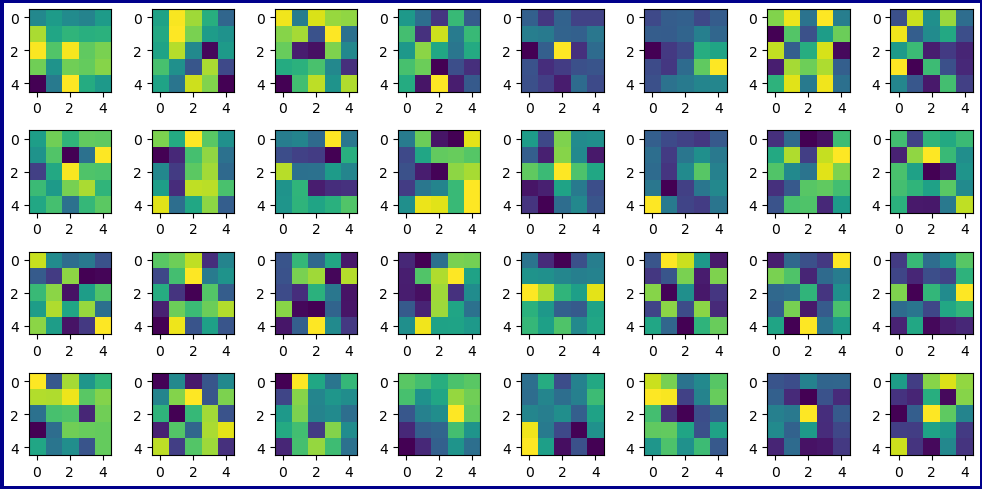}}
\hfill
\subfloat[Block \texttt{\#}5]{\includegraphics[width=0.8\textwidth]{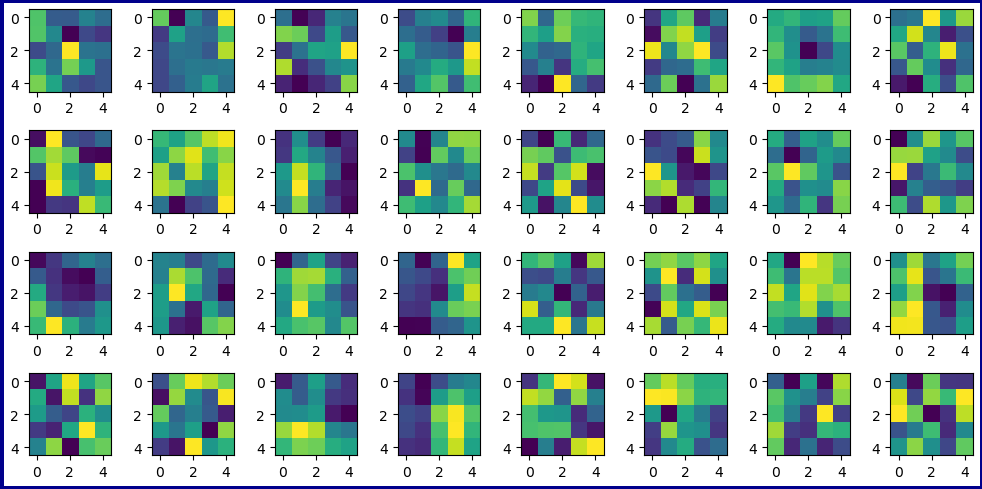}}
\hfill
\subfloat[Block \texttt{\#}6]{\includegraphics[width=0.8\textwidth]{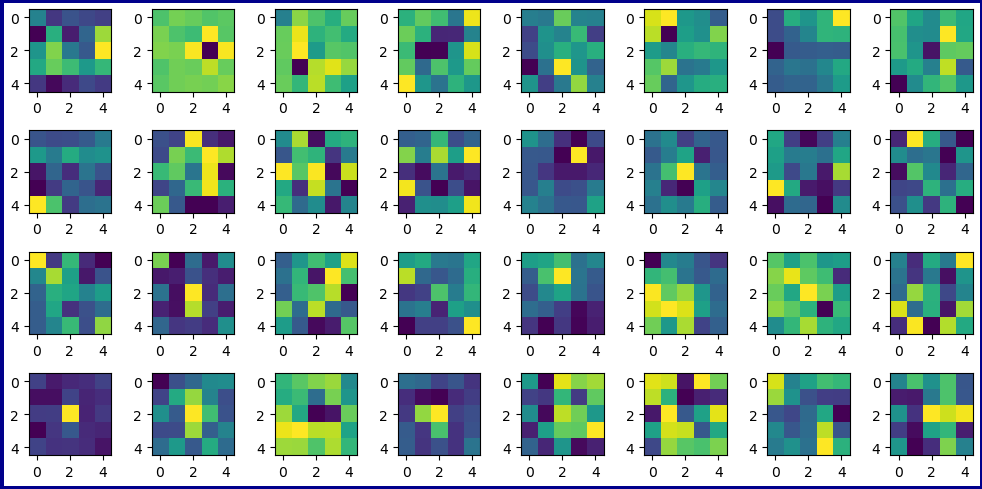}}
	
\caption{(continued)
Some of the kernels of the first convolutional layer for different Blocks (normalized to vary between 0 and 1).
}
\end{figure}
\begin{figure}[!ht]
\captionsetup{belowskip=10pt}
\ContinuedFloat
\centering
\subfloat[Block \texttt{\#}7]{\includegraphics[width=0.8\textwidth]{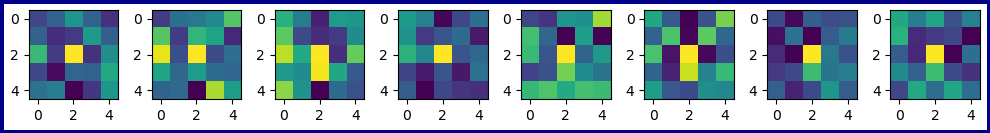}}
\caption{(continued)
Some of the kernels of the first convolutional layer for different Blocks (normalized to vary between 0 and 1).
}
\end{figure}
The kernels of the earlier blocks act as a collection of various liquid-gas interface detectors.
As moving higher, the kernels act more like a set of detectors for various local droplet patterns.

\subsection{Custom image}
\label{custom_image}
\begin{figure}[tb!]
\centering
\includegraphics[width=0.5\textwidth]{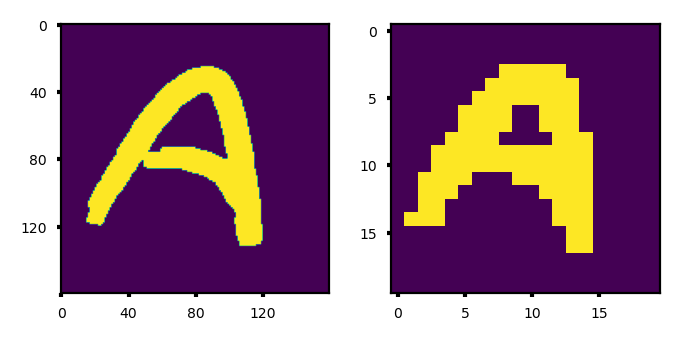}
\caption{
(Left) A custom image of the letter \textit{`A'} together with (Right) the droplet pattern produced by a crude model consisting of three max pooling layers with no trainable parameters.
}
\label{fig_custom_img_letter_A}
\end{figure}

In order to examine how the model performs on a custom image, we used an image of the letter `A' as shown in Fig.~\ref{fig_custom_img_letter_A} (Left).
A crude model consisting of three max pooling layers with no trainable parameters predicts a droplet pattern 
with 97 `On' pixels (number of droplets to be dispensed)
as shown in Fig.~\ref{fig_custom_img_letter_A} (Right).

\begin{figure}[tb!]
\captionsetup{belowskip=10pt}
\centering
\subfloat[]{\includegraphics[width=0.4\textwidth]{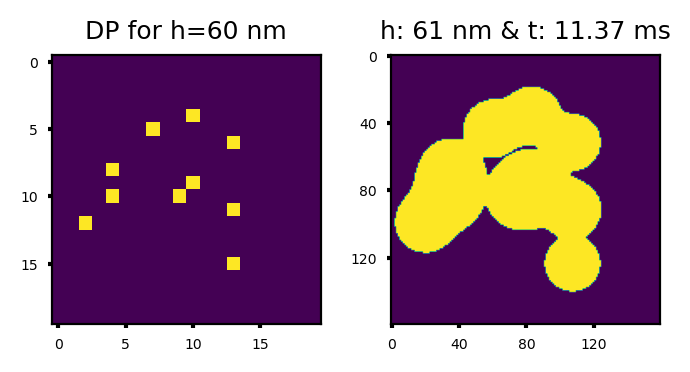}}\hfill
\subfloat[]{\includegraphics[width=0.4\textwidth]{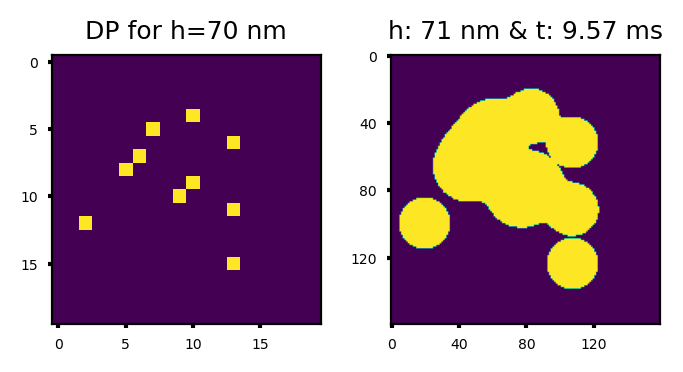}}\hfill
\subfloat[]{\includegraphics[width=0.4\textwidth]{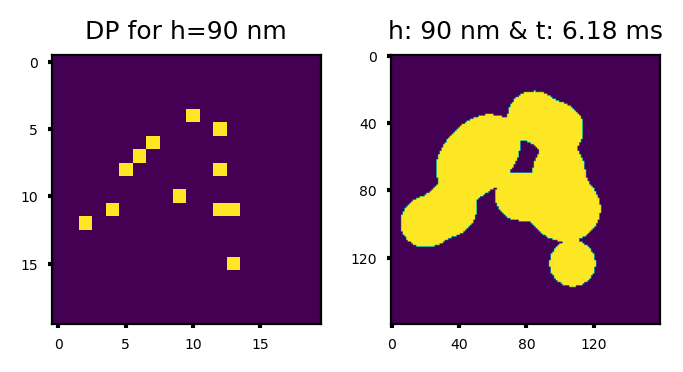}}\hfill
\subfloat[]{\includegraphics[width=0.4\textwidth]{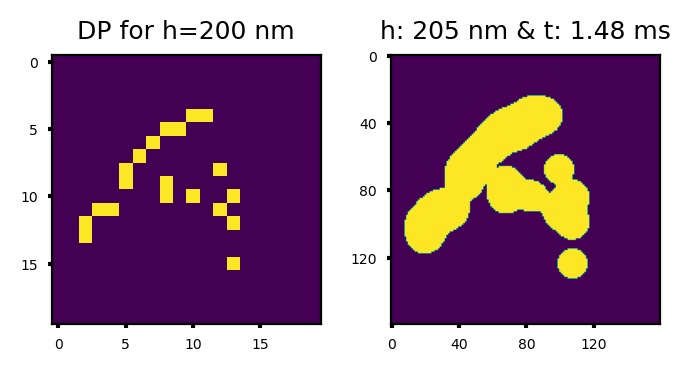}}\hfill

\caption{
(Left) The droplet pattern image (DP) predicted by the CNN-TTR model for the custom image of the letter \textit{`A'} shown in Fig.~\ref{fig_custom_img_letter_A} together with the (Right) corresponding imprint image obtained from the physics-based solver using the predicted droplet pattern image 
for different cases wherein the target liquid film thickness equals
(a) 60 nm, \textit{i.e.} $h^{\ast}=-0.641$, 
(b) 70 nm, \textit{i.e.} $h^{\ast}=-0.469$, 
(c) 90 nm, \textit{i.e.} $h^{\ast}=-0.189$, and
(d) 200 nm, \textit{i.e.} $h^{\ast}=0.701$.
}
\label{fig_custom_img_letter_A_pred}
\end{figure}

We used the CNN-TTR model to predict the droplet pattern images for different target liquid film thicknesses (classification threshold of 0.3). 
Some of the results are shown in Fig.~\ref{fig_custom_img_letter_A_pred}.
It can be perceived that the model generally predicts a fewer number of droplets (`On' pixels in \textit{dp} image) to be dispensed for thinner liquid films (10\textendash20 droplets) 
since the droplets spread more while the liquid film thickness reduces to a smaller value.

In the next step, we examined the droplet patterns predicted for these inverse problems, \textit{i.e.} left images in Fig.~\ref{fig_custom_img_letter_A_pred} (a\textendash d).
We used the physics-based solver to find the imprint \textit{vof} image for the predicted droplet pattern and the corresponding target liquid film thickness: right images in Fig.~\ref{fig_custom_img_letter_A_pred} (a\textendash d).

We notice
that the obtainable imprint images do not have as high resolution as the input image of the letter `A' shown in Fig.~\ref{fig_custom_img_letter_A}.
This can be attributed to the finite size of droplets (6 pl in this work) and the fact that
they become even larger (lower obtainable resolution) as the liquid film is squeezed further.
As a result, the best obtainable resolution is dictated by the size of droplets dispensed by the printhead.

\begin{figure}[tb!]
\centering
\includegraphics[width=0.8\textwidth]{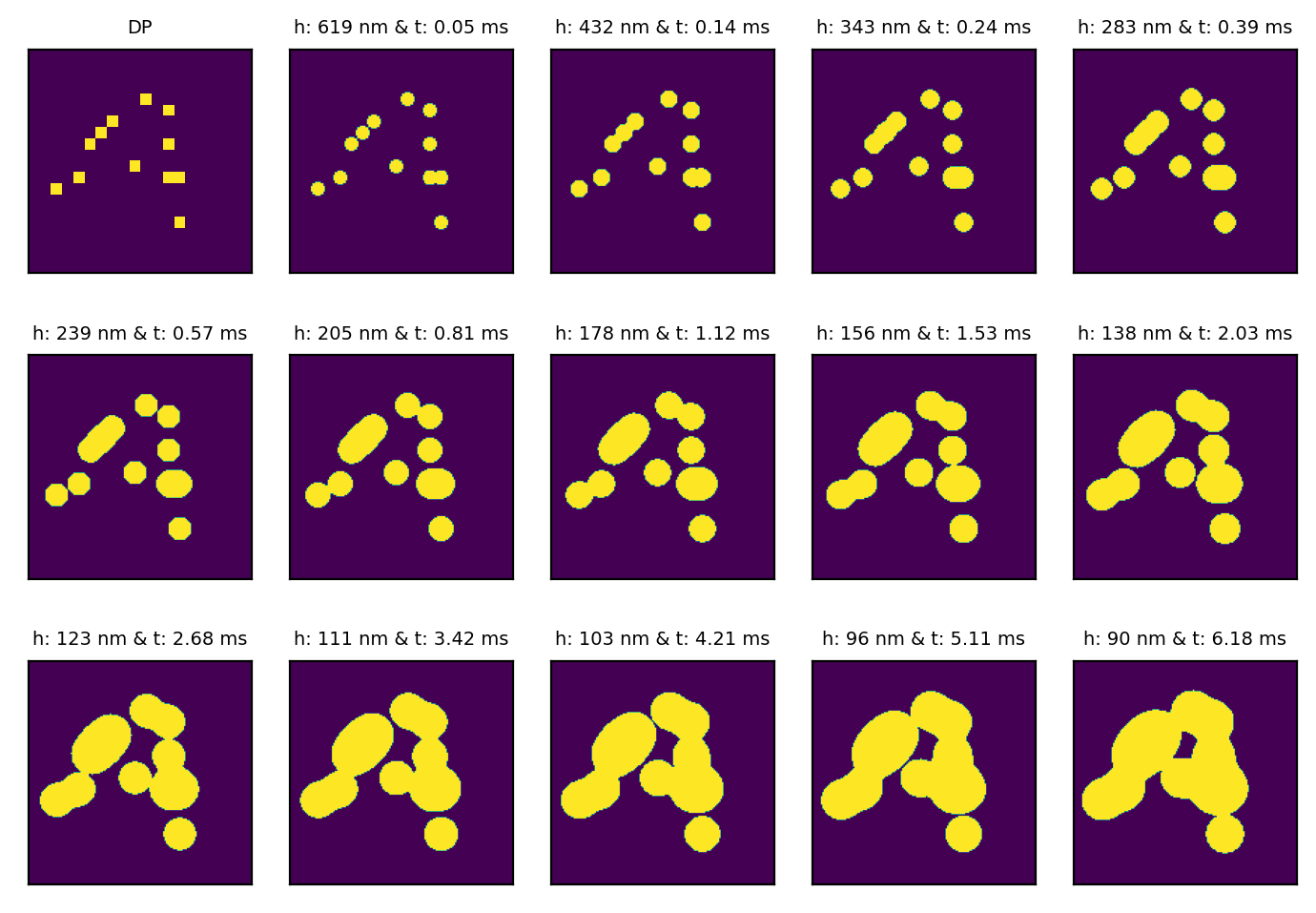}
\caption{
The droplet pattern produced by the CNN-TTR model for a custom image of the letter \textit{`A'} shown in Fig.~\ref{fig_custom_img_letter_A} 
and the target liquid film thickness of 90 nm, \textit{i.e.} $h^{\ast}=-0.189$, 
together with the time series imprint \textit{vof} images obtained from the physics-based solver for the predicted droplet pattern image.
}
\label{fig_custom_img_letter_A_h_90nm_time_series}
\end{figure}

\begin{figure}[tb!]
\centering
\includegraphics[width=0.8\textwidth]{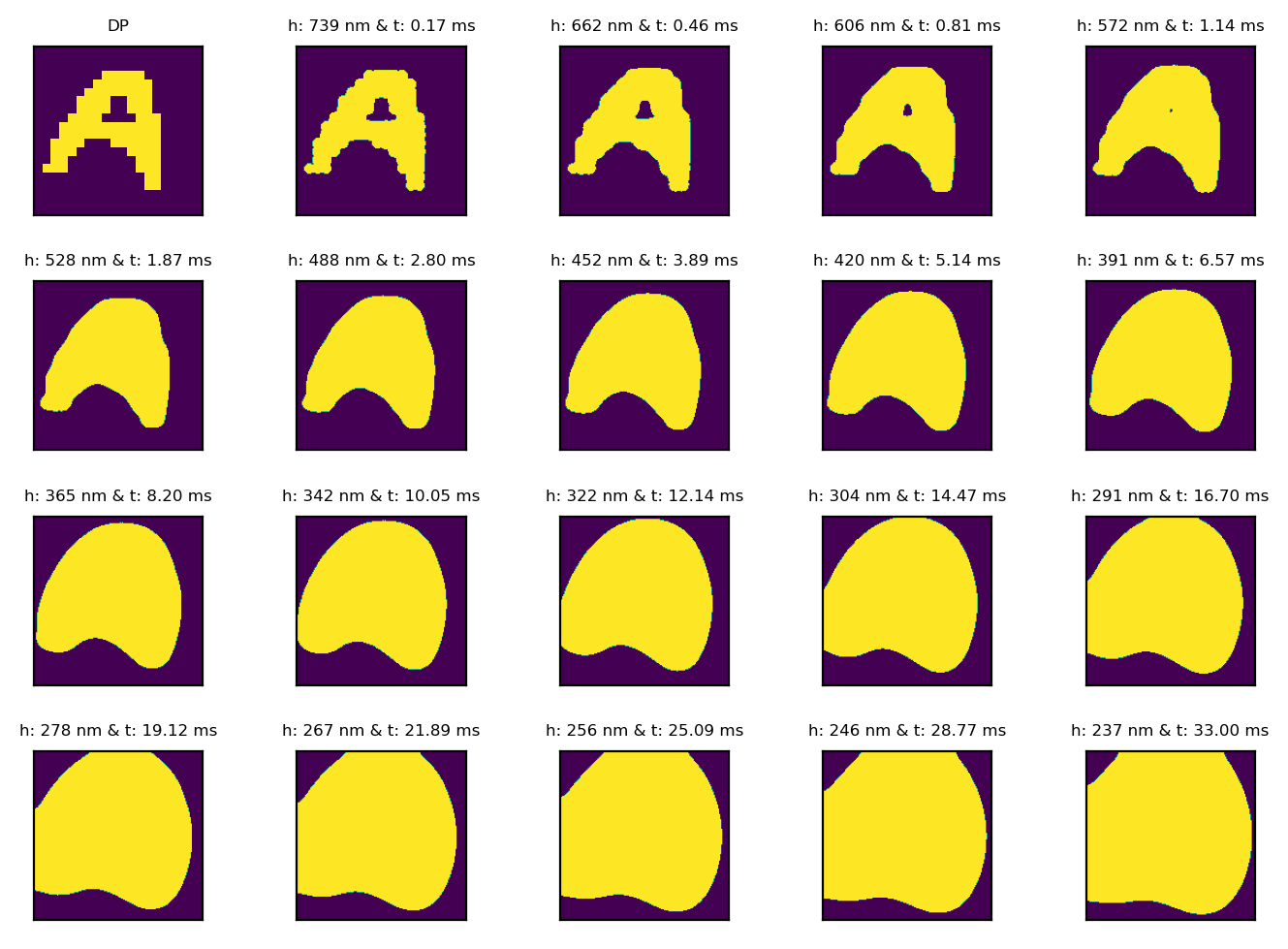}
\caption{
The droplet pattern produced by a crude model consisting of three max pooling layers with no trainable parameters for a custom image of the letter \textit{`A'} shown in Fig.~\ref{fig_custom_img_letter_A} 
together with the time series imprint \textit{vof} images obtained from the physics-based solver for the predicted droplet pattern image.
}
\label{fig_custom_img_letter_A_crude_model_time_series}
\end{figure}

The time series imprint \textit{vof} images obtained by using the physics-based solver for the case of the target liquid film thickness of 90 nm are shown in Fig.~\ref{fig_custom_img_letter_A_h_90nm_time_series}.
The corresponding results related to the droplet pattern predicted by a crude model consisting of three max pooling layers with no trainable parameters are also shown in Fig.~\ref{fig_custom_img_letter_A_crude_model_time_series}.
We can see that the crude model's performance deteriorates as the liquid film thickness decreases since it does not take into account the spread and merge of droplets.

\newpage
\section{Future works}
\label{future works}
The results of the inverse problems shown in Fig.~\ref{fig_custom_img_letter_A_pred} can be improved by human/physics-based solver intervention, \textit{i.e.} using the model's predicted droplet pattern image as a starting point to add, remove, or displace droplets for a more desired outcome.
Hence, the future efforts can be devoted to two purposes: 1. improving the neural network model, and 2. enhancing the neural network model by including the physics-based solver in the decision making process.
Therefore, some of the potential future works could include:

\begin{itemize}
\item including both neural network (inverse problem) and physics-based solver (forward problem) in the decision making process.
\item extending the dataset to include smaller droplets that can potentially lead to higher resolution patterns.
\item extending the dataset to include various droplet diameters to increase the flexibility of generating droplet patterns for higher resolution patterns.
\end{itemize}

\addcontentsline{toc}{section}{Conclusion}
\section*{Conclusion}
\label{conclusion}
In this paper, we proposed a residual convolutional neural network with a new structure suitable for refining a high resolution image to a low resolution one.
We demonstrated the concept in the context of \textit{squeeze flow of micro-droplets}.
We showed that the CNN-TTR model learns to systematically tune the refinement level of 
its residual convolutional blocks by using multiple function approximators.
In future, the datasets can be extended to include various droplet volumes 
to enhance the flexibility of generating droplet pattern maps for producing higher resolution imprint patterns.
We foresee that the proposed platform will be used in data compression and data encryption.
Moreover, it may pave the way for a light-weight LR data communication which can be decoded to its corresponding HR format by using 
the corresponding forward model on each side of the communication.
In addition, the developed package and compiled datasets can potentially 
serve as free, flexible, and scalable standardized benchmarks in connection with
time series images,
multi-class multi-label classifications with many classes/labels, 
image deblurring, etc.

\newpage

\section*{Data Availability Statement}
The developed package and datasets are publicly available on GitHub at \url{https://github.com/sqflow/sqflow}.

\section*{Acknowledgments}
The authors acknowledge the Texas Advanced Computing Center (TACC) at The University of Texas at Austin for providing HPC resources that have contributed to the research results reported within this paper. URL: \url{http://www.tacc.utexas.edu}

\section*{Conflict of Interest}
The authors have no conflicts to disclose.

\bibliographystyle{utphys} 
\bibliography{NIL_ML}

\providecommand{\href}[2]{#2}\begingroup\raggedright\begin{thebibliography}{10}

\bibitem{sreenivasan_nanoimprint_2017}
S.~Sreenivasan, ``Nanoimprint lithography steppers for volume fabrication of
  leading-edge semiconductor integrated circuits,''
  \href{http://dx.doi.org/10.1038/micronano.2017.75}{{\em Microsystems \&
  Nanoengineering} {\bfseries 3} no.~1, (Dec., 2017) }.
  \url{http://www.nature.com/articles/micronano201775}.

\bibitem{reddy_simulation_2005}
S.~Reddy and R.~T. Bonnecaze,
  \href{http://dx.doi.org/10.1117/12.601413}{``Simulation of fluid flow in the
  step and flash imprint lithography process,''} p.~200.
\newblock San Jose, CA, May, 2005.
\newblock
  \url{http://proceedings.spiedigitallibrary.org/proceeding.aspx?doi=10.1117/12.601413}.

\bibitem{pang_image--image_2021}
Y.~Pang, J.~Lin, T.~Qin, and Z.~Chen, ``Image-to-{Image} {Translation}:
  {Methods} and {Applications},'' July, 2021.
\newblock \url{http://arxiv.org/abs/2101.08629}. arXiv:2101.08629 [cs].

\bibitem{lanchantin_general_2020}
J.~Lanchantin, T.~Wang, V.~Ordonez, and Y.~Qi, ``General {Multi}-label {Image}
  {Classification} with {Transformers},'' Nov., 2020.
\newblock \url{http://arxiv.org/abs/2011.14027}. arXiv:2011.14027 [cs].

\bibitem{yun_re-labeling_2021}
S.~Yun, S.~J. Oh, B.~Heo, D.~Han, J.~Choe, and S.~Chun, ``Re-labeling
  {ImageNet}: from {Single} to {Multi}-{Labels}, from {Global} to {Localized}
  {Labels},'' July, 2021.
\newblock \url{http://arxiv.org/abs/2101.05022}. arXiv:2101.05022 [cs].

\bibitem{liu_emerging_2022}
W.~Liu, H.~Wang, X.~Shen, and I.~W. Tsang, ``The {Emerging} {Trends} of
  {Multi}-{Label} {Learning},''
  \href{http://dx.doi.org/10.1109/TPAMI.2021.3119334}{{\em IEEE Transactions on
  Pattern Analysis and Machine Intelligence} {\bfseries 44} no.~11, (Nov.,
  2022) 7955--7974}. \url{http://arxiv.org/abs/2011.11197}. arXiv:2011.11197
  [cs].

\bibitem{wang_deep_2020}
Z.~Wang, J.~Chen, and S.~C.~H. Hoi, ``Deep {Learning} for {Image}
  {Super}-resolution: {A} {Survey},'' Feb., 2020.
\newblock \url{http://arxiv.org/abs/1902.06068}. arXiv:1902.06068 [cs].

\bibitem{li_beginner_2021}
J.~Li, Z.~Pei, and T.~Zeng, ``From {Beginner} to {Master}: {A} {Survey} for
  {Deep} {Learning}-based {Single}-{Image} {Super}-{Resolution},'' Sept., 2021.
\newblock \url{http://arxiv.org/abs/2109.14335}. arXiv:2109.14335 [cs, eess].

\bibitem{liu_blind_2021}
A.~Liu, Y.~Liu, J.~Gu, Y.~Qiao, and C.~Dong, ``Blind {Image}
  {Super}-{Resolution}: {A} {Survey} and {Beyond},'' July, 2021.
\newblock \url{http://arxiv.org/abs/2107.03055}. arXiv:2107.03055 [cs].

\bibitem{saharia_image_2021}
C.~Saharia, J.~Ho, W.~Chan, T.~Salimans, D.~J. Fleet, and M.~Norouzi, ``Image
  {Super}-{Resolution} via {Iterative} {Refinement},'' June, 2021.
\newblock \url{http://arxiv.org/abs/2104.07636}. arXiv:2104.07636 [cs, eess].

\bibitem{maral_single_2022}
B.~C. Maral, ``Single {Image} {Super}-{Resolution} {Methods}: {A} {Survey},''
  Feb., 2022.
\newblock \url{http://arxiv.org/abs/2202.11763}. arXiv:2202.11763 [cs, eess].

\bibitem{zhao_new_2020}
H.~Zhao, Z.~Ke, N.~Chen, S.~Wang, K.~Li, L.~Wang, X.~Gong, W.~Zheng, L.~Song,
  Z.~Liu, D.~Liang, and C.~Liu, ``A new deep learning method for image
  deblurring in optical microscopic systems,''
  \href{http://dx.doi.org/10.1002/jbio.201960147}{{\em Journal of Biophotonics}
  {\bfseries 13} no.~3, (Mar., 2020) }.
  \url{https://onlinelibrary.wiley.com/doi/10.1002/jbio.201960147}.

\bibitem{tao_scale-recurrent_nodate}
X.~Tao, H.~Gao, X.~Shen, J.~Wang, and J.~Jia, ``Scale-{Recurrent} {Network} for
  {Deep} {Image} {Deblurring},''.

\bibitem{zhang_deep_2022}
K.~Zhang, W.~Ren, W.~Luo, W.-S. Lai, B.~Stenger, M.-H. Yang, and H.~Li, ``Deep
  {Image} {Deblurring}: {A} {Survey},'' May, 2022.
\newblock \url{http://arxiv.org/abs/2201.10700}. arXiv:2201.10700 [cs].

\bibitem{hamrock_fundamentals_1991}
B.~Hamrock, {\em Fundamentals of {Fluid} {Film} {Lubrication}}.
\newblock NASA Reference Publication 1255, 1991.

\bibitem{mehboudi_modeling_2022}
A.~Mehboudi, S.~Singhal, and S.~V. Sreenivasan, ``Modeling the squeeze flow of
  droplet over a step,'' \href{http://dx.doi.org/10.1063/5.0098597}{{\em
  Physics of Fluids} {\bfseries 34} no.~8, (Aug., 2022) 082005}.
  \url{https://aip.scitation.org/doi/10.1063/5.0098597}. Publisher: American
  Institute of Physics.

\bibitem{hornik_multilayer_1989}
K.~Hornik, M.~Stinchcombe, and H.~White, ``Multilayer feedforward networks are
  universal approximators,''
  \href{http://dx.doi.org/10.1016/0893-6080(89)90020-8}{{\em Neural Networks}
  {\bfseries 2} no.~5, (Jan., 1989) 359--366}.
  \url{https://linkinghub.elsevier.com/retrieve/pii/0893608089900208}.

\bibitem{glorot_deep_2011}
X.~Glorot, A.~Bordes, and Y.~Bengio, ``Deep {Sparse} {Rectiﬁer} {Neural}
  {Networks},'' {\em Artificial Intelligence and Statistics} {\bfseries 15}
  (2011) 9.

\bibitem{ulyanov_instance_2017}
D.~Ulyanov, A.~Vedaldi, and V.~Lempitsky, ``Instance {Normalization}: {The}
  {Missing} {Ingredient} for {Fast} {Stylization},'' Nov., 2017.
\newblock \url{http://arxiv.org/abs/1607.08022}. arXiv:1607.08022 [cs].

\bibitem{abadi_tensorflow_2016}
M.~Abadi, A.~Agarwal, P.~Barham, E.~Brevdo, Z.~Chen, C.~Citro, G.~S. Corrado,
  A.~Davis, J.~Dean, M.~Devin, S.~Ghemawat, I.~Goodfellow, A.~Harp, G.~Irving,
  M.~Isard, Y.~Jia, R.~Jozefowicz, L.~Kaiser, M.~Kudlur, J.~Levenberg, D.~Mane,
  R.~Monga, S.~Moore, D.~Murray, C.~Olah, M.~Schuster, J.~Shlens, B.~Steiner,
  I.~Sutskever, K.~Talwar, P.~Tucker, V.~Vanhoucke, V.~Vasudevan, F.~Viegas,
  O.~Vinyals, P.~Warden, M.~Wattenberg, M.~Wicke, Y.~Yu, and X.~Zheng,
  ``{TensorFlow}: {Large}-{Scale} {Machine} {Learning} on {Heterogeneous}
  {Distributed} {Systems},'' Mar., 2016.
\newblock \url{http://arxiv.org/abs/1603.04467}. arXiv:1603.04467 [cs].

\bibitem{chollet_keras_2015}
F.~Chollet and {others}, ``Keras,'' 2015.
\newblock \url{https://keras.io}.

\bibitem{saito_precision-recall_2015}
T.~Saito and M.~Rehmsmeier, ``The {Precision}-{Recall} {Plot} {Is} {More}
  {Informative} than the {ROC} {Plot} {When} {Evaluating} {Binary}
  {Classifiers} on {Imbalanced} {Datasets},''
  \href{http://dx.doi.org/10.1371/journal.pone.0118432}{{\em PLOS ONE}
  {\bfseries 10} no.~3, (Mar., 2015) e0118432}.
  \url{https://dx.plos.org/10.1371/journal.pone.0118432}.

\end{thebibliography}\endgroup

\end{document}